\documentclass[journal]{IEEEtran}
\usepackage{amsmath,epsfig}
\usepackage{tikz}

\usepackage{float}
\usepackage{xfrac}
\usepackage{amsfonts}
\usepackage{cite}
\usepackage{algorithm}
\usepackage{algpseudocode}
\usepackage{tabularx}
\usepackage{graphicx}
\usepackage{slashbox}
\usepackage{epstopdf}
\usepackage{color}
\usepackage{subfigure}
\usepackage{multitoc}
\usepackage{caption}
\usepackage{color}

\usepackage{makecell}
\newcolumntype{x}[1]{>{\centering\let\newline\\\arraybackslash\hspace{0pt}}p{#1}}

\def\NoNumber#1{{\def\alglinenumber##1{}\State #1}\addtocounter{ALG@line}{-1}}
\def\balpha{\boldsymbol\alpha}
\def\bGamma{\boldsymbol\Gamma}

\def\bh{\mathbf{h}}
\def\bx{\mathbf{x}}
\def\bE{\mathbf{E}}
\def\bX{\mathbf{X}}

\def\bD{\mathbf{D}}

\def\bB{\mathbf{B}}
\def\bW{\mathbf{W}}
\def\by{\mathbf{y}}
\def\bY{\mathbf{Y}}
\def\bV{\mathbf{V}}
\def\bF{\mathbf{F}}
\def\bg{\mathbf{g}}
\def\bI{\mathbf{I}}
\def\bC{\mathbf{C}}
\def\bA{\mathbf{A}}
\def\bZ{\mathbf{Z}}
\def\bU{\mathbf{U}}

\def\bz{\mathbf{z}}

\def\bbR{{\mathbb R}}

\def\calL{{\cal L}}
\def\calJ{{\cal J}}
\def\bL{{\mathbf L}}
\def\tbD{\tilde{\mathbf D}}
\def\tD{\tilde{D}}
\def\tbA{\tilde{\mathbf A}}
\def\tbE{\tilde{\mathbf E}}
\def\tbW{\tilde{\mathbf W}}
\def\bE{\mathbf{E}}

\def\bgamma{\boldsymbol\gamma}
\def\bxi{\boldsymbol\xi}
\def\bGamma{\boldsymbol\Gamma}

\def\bbeta{\boldsymbol\beta}
\makeatletter
\newcommand*{\rom}[1]{\expandafter\@slowromancap\romannumeral #1@}
\makeatother

 \newcommand{\norm}[1]{\lVert #1 \rVert}

\makeatother

\begin{document}
%\setpagewiselinenumbers
%\modulolinenumbers[1]
%\linenumbers

\title{Task-Driven Dictionary Learning for Hyperspectral Image Classification with Structured Sparsity Constraints}
%
% Single address.
% ---------------
%\name{ Xiaoxia Sun$^\dag$, Nasser M. Nasrabadi$^\ddag$ and Trac D. Tran$^\dag$ \thanks{This work has
%been partially supported by NSF (National Science Foundation) under Grants CCF-1117545, ARO (Army Research Office) under
%Grants 60219-MA, and ONR (Office of Naval Research) under grant N000141210765.}}
%\address{ $\dag$ The Johns Hopkins University, Baltimore, MD 21218
%USA  (xsun9@jhu.edu)\\
%$\ddag$ U.S. Army Research Laboratory, Adelphi, MD
%20783 USA}
\author{Xiaoxia Sun,  Nasser M. Nasrabadi,~\IEEEmembership{Fellow,~IEEE,} and Trac D. Tran,~\IEEEmembership{Fellow,~IEEE}
\thanks{X.Sun and T. D. Tran are with the Department of Electrical and Computer Engineering, The Johns Hopkins University, Baltimore, MD 21218 USA (e-mail: xsun9@jhu.edu; trac@jhu.edu). This work has been partially supported by NSF under Grants CCF-1117545, ARO under Grants 60219-MA, and ONR under grant N000141210765.} 
\thanks{N. M. Nasrabadi is with U.S. Army Research Laboratory, Adelphi, MD 20783
USA (e-mail: nnasraba@arl.army.mil).} }

\markboth{IEEE Transactions on Geoscience and Remote Sensing}
{Task-driven Dictionary Learning for Hyperspectral Image Classification with Structured Sparsity Priors}

%\ninept
%
\maketitle
%
%exploiting the spatial-spectral dependencies of neighboring pixels and 
\begin{abstract}
Sparse representation models a signal as a linear combination of a small number of dictionary atoms. As a generative model, it requires the dictionary to be highly redundant in order to ensure both a stable high sparsity level and a low reconstruction error for the signal. However, in practice, this requirement is usually impaired by the lack of  labelled training samples.   Fortunately, previous research has shown that the requirement for a redundant dictionary can be less rigorous if simultaneous sparse approximation is employed, which can be carried out by enforcing various structured sparsity constraints on the sparse codes of the neighboring pixels. In addition, numerous works have shown that applying a variety of dictionary learning methods for the sparse representation model can also  improve the classification performance. In this paper, we highlight the task-driven dictionary learning algorithm, which is a general framework for the supervised dictionary learning method.  We propose to enforce  structured sparsity priors on the  task-driven dictionary learning method in order  to improve the performance of the hyperspectral classification. Our approach is able to benefit from both the advantages of the  simultaneous sparse representation and those of the supervised dictionary learning.  We enforce two different structured sparsity priors, the joint and Laplacian sparsity, on the task-driven dictionary learning method and provide the details of the corresponding optimization algorithms. Experiments on numerous popular hyperspectral images demonstrate that the classification performance of our approach is superior to sparse representation classifier with structured priors or the task-driven dictionary learning method.  
%The proposed method has numerous advantages over the existing dictionary learning techniques. %  and Laplacian sparsity prior.  
% It uses a structured sparsity and provides a more robust and stable sparse coefficients. Besides,  it is capable of reducing the classification error by jointly optimizing the dictionary and the classifier's parameters during the dictionary training stage. Experiments on numerous popular hyperspectral images demonstrate that our approach is superior to the existing dictionary learning methods.
% In this paper, we first discuss dictionary learning with joint sparsity. In order to improve the performance on non-homogeneous regions, we will propose dictionary learning with Laplacian sparsity in the second part of this paper.
% Moreover, the proposed method can significantly reduce the dictionary size while maintaining a plausible classification performance. 
\end{abstract}
\section{Introduction}
\label{sec:intro}

\PARstart{C}{lassification} on Hyperspectral Imagery (HSI) is becoming increasingly popular in remote sensing. Notable applications include military aerial surveillance \cite{NasserSPM,Chen_target,hsi_application}, mineral identification  and material defects detection \cite{advances_SPM}. However, numerous difficulties impede the improvement of HSI classification performance. For instance, the high dimensionality of HSI pixels introduce the problem of the `curse of dimensionality' \cite{curse_dim}, and the classifier is always confronted with the overfitting problem due to the small number of  labelled samples. Additionally, most HSI pixels are indiscriminative since they are undesirably highly coherent \cite{sparse_unx}. In the past few decades, numerous classification techniques, such as SVM \cite{Melgani}, k-nearest-neighbor classifier \cite{knearest_neighbors}, multimodel logistic regression \cite{logistic_regression} and neural network \cite{Benediktsson}, have been proposed to alleviate some of these problems to achieve an acceptable performance for HSI classification. 
 
 \subsection{Sparse Representation for HSI classification}
 More recently, researchers have focused attention on describing the high dimensional data as a sparse linear combination of dictionary atoms.  Sparse representation classifier (SRC)   was proposed in \cite{jwright} and has been successfully applied to a wide variety of applications, such as face recognition \cite{jwright}, visual tracking \cite{tracking_src}, speech recognition \cite{speech_recog} and aerial image detection \cite{detection_src}. SRC has also been applied to HSI classification by Chen \emph{et. al.} \cite{Chen}, where a dictionary was constructed by stacking all the labelled samples. Success of SRC requires  that the high dimensional data belonging to the same class to lie in a low dimensional subspace.
% Success of SRC relies on the observation that most of the high dimensional data belonging to the same class usually lie in the same low dimensional subspace.  
  The outstanding classification performance is due to the robustness of sparse recovery, which is largely provided by the high redundancy and low coherency of the dictionary atoms.  A low reconstruction error and a high sparsity level can be achieved if the designed dictionary satisfies the above properties. Unfortunately, in practice,  the HSI dictionary usually does not have the above properties due to the small number of blue{highly correlated} labelled training samples \cite{sparse_unx}. 
 
%Due to the poor dictionary construction, 
Due to these undesired properties of the HSI dictionary, the sparse recovery can become unstable and unpredictable such that even pixels belonging to the same class can have totally different sparse codes. The problem induced by the high-coherency of the dictionary atoms, which can be alleviated through decreasing the variation between the sparse codes of  the hyperspectral pixels that belong to the same class.  In HSI, pixels that are spatially close to each other usually have similar spectral features and belong to the same class. Previous research has shown that the sparse codes of neighboring pixels can become similar by enforcing a structured sparsity constraint (prior). The simultaneous sparse recovery is analytically guaranteed to achieve a sparser solution and a lower reconstruction error with a smaller dictionary \cite{theoretical_mmv}. A variety of structured sparsity priors are proposed in the literature \cite{Xiaoxia} that are capable of generating  different desired sparsity patterns for the sparse codes of neighboring pixels.  The joint sparsity prior \cite{Chen} assumes that the features of all the neighboring pixels lie in the same low dimensional subspace and all the corresponding sparse codes share the same set of dictionary atoms. Therefore, the sparse codes have a row sparsity pattern, where only a few rows of the sparse codes are nonzero \cite{Tropp_js, Cotter_js}. The collaborative group sparsity prior \cite{cgroup} enforces the coefficients to have a  group-wise sparsity pattern, where the coefficients  within each active group are dense. The collaborative hierarchical  sparsity prior \cite{chilasso} enforces the sparse codes to be not only group-wise sparse, but also sparse within each active group. The low rank prior \cite{low_rank} assumes that the neighboring pixels are linearly dependent. It does not necessary lead the coefficients to be sparse, which is detrimental for a good classification. However, the low rank group prior proposed in \cite{Xiaoxia} is able to enforce both a group sparsity prior and a low rank prior on the sparse codes by forcing the same group of dictionary atoms to be active if and only if  the corresponding neighboring pixels are linearly dependent.    The Laplacian sparsity prior \cite{Gao} uses a Laplacian matrix to describe the degree of similarity between the neighboring pixels. The neighboring  pixels  that have less spectral features in common are less encouraged to have a similar sparse codes.  It has been shown that all the  structured sparsity priors are capable of obtaining a smoother classification map and improving the  classification performance  \cite{Xiaoxia}.  

\subsection{Dictionary Learning for Sparse Representation}
In the classical SRC, the dictionary is constructed by stacking all the training samples. The sparse recovery can be computationally burdensome when the training set is large. Besides, the dictionary constructed in this manner can neither be optimal for reconstruction purposes nor for classification of  signals. Previous literature have shown that a dictionary can be trained to have a better representation of the dataset. Unsupervised dictionary learning methods, such as the method of optimal direction (MOD) \cite{mod}, K-SVD  \cite{Aharon} and online dictionary learning \cite{Mairal}, are able to improve the signal restoration performance of numerous applications, such as compressive sensing, signal denoising and image inpainting.

However, the unsupervised dictionary learning method  is not suitable for solving classification problems since a lower reconstruction error does not necessarily lead to a better classification performance. In fact, it is  observed that the dictionary can have an improved classification result by sacrificing some signal reconstruction performance \cite{MairalTDDL}.  Therefore, supervised dictionary learning methods \cite{mairal_sd} are proposed to improve the classification result. Unlike the unsupervised dictionary learning, which only trains the dictionary by pursuing a lower signal reconstruction error, the supervised learning is able to directly improve the classification performance by optimizing both the dictionary and classifier's parameter simultaneously.  The discriminative dictionary learning in \cite{Mairal_dis} minimizes the classification error of SRC by minimizing the reconstruction error contributed by the atoms from the correct class and maximizing the error from the remaining classes.  The incoherent dictionary learning in \cite{incoh_dl} uses SRC as the classifier and tries to  eliminate the atoms shared by pixels from different classes. It increases the discriminability of the sparse codes by decreasing the coherency of the atoms from different classes. The label consistent K-SVD (LC-KSVD) \cite{Jiang} optimizes the dictionary and classifier's parameter by minimizing the summation of reconstruction and classification errors. It combines the dictionary and classifier's parameter into a single parameter space, which makes it possible for the optimization procedure to be much simpler than those used in classical SRC.  However, a desired and accurate solution is not guaranteed \cite{yang_coupled} because the cost function can be minimized by decreasing the reconstruction error while the classification error is increased. A bilevel optimization formulation would be more appropriate \cite{bilevel_survey}, where the update of the dictionary is driven by the minimization of the classification error.  The task-driven dictionary learning (TDDL) \cite{MairalTDDL} exploits this idea with theoretical proof and demonstrates a superior performance.   The supervised translation-invariant sparse coding, which uses the same scheme as TDDL,   is developed independently by \cite{Yang}. It is a more general framework that can be applied not only to classification, but also nonlinear image mapping, digital art authentiﬁcation and compressive sensing. More recently, the group sparsity prior is enforced on a single measurement and  the corresponding TDDL optimization algorithm is developed in  \cite{tag_taxo} in order to improve the performance of region tagging.

\subsection{Contributions}
In this paper, we propose a novel method that enforces the joint or Laplacian sparsity prior on the sparse recovery stage of TDDL.  The existing dictionary learning methods have only been developed for reconstructing or classifying a single measurement.  Therefore, it is advantageous to incorporate structured sparsity priors into the supervised dictionary learning  in order to achieve a better performance. This paper makes the following contributions:
\begin{itemize}
\item We propose a new dictionary learning algorithm for TDDL with joint or Laplacian sparsity in order to exploit the spatial-spectral information of HSI neighboring pixels.
\item We show experimentally that the proposed dictionary learning methods have a significantly better performance than SRC even when the dictionary is highly compact.
\item We also describe an optimization algorithm for solving the Laplacian sparsity recovery problem. The proposed optimization method is much faster than the modified feature sign search used in \cite{Gao}.

\end{itemize}

%Similar as in SRC, the proposed method also benefits from better reconstructed signals and a more robust sparse representation codes. Moreover, during the training stage, the dictionary adjustments depend on both the labeled center pixel and their neighboring pixels. In addition, enforcing the Laplacian prior gives a better performance on the non-homogeneous regions when compared with using the joint sparsity prior.

The remainder of the paper is organized as follows. In Section \ref{sec:related_works}, a brief review of  TDDL is given. In Section \ref{sec:tddl_js}, we propose a modified TDDL algorithm  with the joint sparsity prior. TDDL with the Laplacian prior and a new algorithm for recovering the Laplacian sparse problem are stated in Section \ref{sec:laplacian}. In Section \ref{sec:results}, we show that our method is superior to other HSI classification methods through experimental results on several HSI images. Finally, we provide our conclusion in Section \ref{sec:conclusion}.

\section{Task-driven Dictionary Learning}
\label{sec:related_works}

In TDDL \cite{MairalTDDL}, signals are represented by their sparse codes, which are then fed into a linear regression or logistic regression. Consider a pair of training samples $(\bx, \by)$, where $\bx \in \bbR ^{M}$ is the HSI pixel, $M$ is the number of spectral bands,  and $\by \in \bbR^K$ is a binary vector representation of the label of the sample $\bx$. $K$ is the maximum class index. Pixel $\bx$ can be represented by a sparse coefficient vector $\balpha(\bD, \bx)\in \bbR^N$ with respect to some dictionary $\bD \in \bbR^{M\times N}$ consisting of $N$ atoms by solving the optimization
\begin{equation}
\label{eq:elastic_net}
\balpha(\bD, \bx) = \arg\min_{\bz} \lVert \bx - \bD\bz \rVert_2^2 + \lambda\lVert \bz \rVert_1 +\frac{\epsilon}{2}\lVert \bz \rVert_2^2,
\end{equation},where $\lambda$ and $\epsilon$ are the regularization parameters. $\lambda$ controls the sparsity level of the coefficients $\balpha$. In our experiments, we set $\epsilon$ to $0$ since it does not affect the convergence of the algorithm and always gives satisfactory results. 

%Given an arbitrary dictionary $\bD$ consisting of $N$ training samples (atoms), such as the one used in SRC, it is not optimal and can have high possibility for misclassification.

 To optimize the dictionary, TDDL first defines a convex function $\calL( \bD, \bW, \{\bx_i\}_{i=1}^S )$ to describe the classification risk in terms of the dictionary atoms, sparse coefficients and the classifier's parameter $\bW$. The function is then minimized as follows
\begin{equation}
\min_{\bD, \bW} \calL ( \bD, \bW, \{\bx_i\}_{i=1}^S ) =  \min_{\bD, \bW} f(\bD, \bW, \{\bx_i\}_{i=1}^S) + \frac{\mu}{2}\lVert \bW \rVert_F^2,
\end{equation}
where  $\mu>0$ is a classifier regularization parameter to avoid overfitting of the classifier \cite{intro_mh}.
The convex function $f$ is defined as
\begin{equation}
f(\bD, \bW, \{\bx_i\}_{i=1}^S) \overset{\Delta}{=} \frac{1}{S}\sum_{i=1}^S \calJ(\by_i, \bW, \balpha_i(\bD, \bx_i)),
\end{equation}
where $S$ is the total number of training samples and $\calL(\by_i, \bW, \balpha_i(\bD, \bx_i)) $ is the classification error for a training pair $(\bx_i, \by_i)$ which  is measured by a linear regression, i.e. $\calJ(\by_i, \bW, \balpha_i(\bD, \bx_i)) = \frac{1}{2}\lVert \by_i - \bW \balpha_i \rVert_2^2$.

In the following part of the section, we omit the subscript $i$ of $\balpha$ for notational simplicity. The dictionary $\bD$ and the classifier parameter $\bW$ are updated using a  stochastic gradient descent algorithm, which has been independently investigated by \cite{MairalTDDL,Yang}. The update rules for $\bD$ and $\bW$ are
%\begin{equation}
% \begin{cases}
%    \bD^{(t+1)} = \bD^{(t)} - \rho^{(t)} \nabla_{\bD} \calL(\by, \bW^{(t)}, \balpha(\bD^{(t)}, \bx)),\\
%    \bW^{(t+1)} = \bW^{(t)} - \rho^{(t)} \nabla_{\bW} \calL(\by, \bW^{(t)}, \balpha(\bD^{(t)}, \bx)),
%  \end{cases}
%\end{equation}
\begin{equation}
 \begin{cases}
    \bD^{(t+1)} = \bD^{(t)} - \rho^{(t)} \cdot {\partial \calL^{(t)}} / {\partial \bD},\\
    \bW^{(t+1)} = \bW^{(t)} - \rho^{(t)} \cdot {\partial\calL^{(t)}}/{\partial\bW},
  \end{cases}
\end{equation}
where $t$ is the iteration index and $\rho$ is the step size.  The equations for updating the classifier parameter $\bW$ is straightforward since $\calL(\by, \bW, \balpha(\bD, \bx))$  is both smooth and convex with respect to  $\bW$. We have
\begin{equation}
\frac{\partial \calL}{\partial \bW} =  \left( \bW \balpha - \by \right)\balpha^\top + \mu \bW.
\end{equation}
 The updating equation for the dictionary can be obtained by applying error backpropagation, where the chain rule is applied
\begin{equation}
\frac{\partial \calL}{\partial \bD} =  \frac{\partial \calL}{\partial \balpha}\frac{\partial \balpha}{\partial \bD}.
\label{eq:chain_rule}
\end{equation}

The difficulty of acquiring a specific form of the above equation comes from $\sfrac{\partial \balpha}{\partial \bD}$. Since the sparse coefficient $\balpha(\bD, \bx)$ is an implicit function of $\bD$, an analytic form of $\balpha$ with respect to $\bD$ is not available. Fortunately, the derivative $\sfrac{\partial \balpha}{\partial \bD}$ can still be computed by either applying optimality condition of elastic net \cite{MairalTDDL,optElasticNet} or using fixed point differentiation \cite{Yang,differentiablesparse}.

We now focus on computing the derivative using the fixed point differentiation. As suggested in \cite{differentiablesparse}, the gradient of Eq. (\ref{eq:elastic_net}) reaches $\mathbf{0}$ at the optimal point $\hat{\balpha}$
\begin{equation}
\frac{\partial   \lVert  \bx -\bD\balpha \rVert_2^2}{\partial \balpha} \Big|_{\balpha = \hat{\balpha}} = - \lambda  \frac{\partial \lVert \balpha \rVert_1}{\partial \balpha} \Big|_{\balpha = \hat{\balpha}}.
\label{eq:fixedpoint1}
\end{equation}
Expanding Eq. (\ref{eq:fixedpoint1}), we have
\begin{equation}
2\bD^\top (\bx - \bD \balpha) \Big|_{\balpha = \hat{\balpha}} = 
 \lambda \cdot sign(\balpha) \Big|_{\balpha = \hat{\balpha}}.
\label{eq:sign}
\end{equation}

In order to evaluate $\sfrac{\partial \balpha}{\partial \bD}$, the derivative of Eq. (\ref{eq:sign}) with respect to each element $D_{mn}$ of the dictionary is required. Since the differentiation of the $sign$ function is not well defined at zero points, we can only compute the derivative of Eq. (\ref{eq:sign}) at fixed points when $\balpha_{[n]} \neq 0$ \cite{Yang}

\begin{align}
 \frac{\partial \balpha_\Lambda}{\partial D_{mn}} =  (\bD_\Lambda^\top \bD_\Lambda)^{-1}\left( \frac{\partial \bD_\Lambda^\top \bx}{\partial D_{mn}} - \frac{\partial \bD_\Lambda^\top \bD_\Lambda}{\partial D_{mn}}\balpha_\Lambda \right) \mbox{ and  }  \frac{\partial \balpha_{\Lambda^C}}{\partial D_{mn}} = \mathbf{0} ,
\end{align}
where $\Lambda$ and $\Lambda^c$ are the indices of the active and inactive set of $\balpha$ respectively. $D_{mn}\in\bbR$ is the $(m,n)$ element of $\bD$. $(\bD_\Lambda^\top \bD_\Lambda)^{-1}$ is always invertible  since the number of active atoms $\lvert \Lambda\rvert$ is always much smaller than the feature dimension $M$. 

%The element-wise updating of $\bD$ can be further simplified to a computational friendly form as the one discussed in \cite{MairalTDDL}, which is derived from the optimality condition of the elastic net. In the following sections, we develop the algorithm of TDDL with structured sparsity priors using fixed point method and simplify the algorithm into a more concise form.

\section{TDDL WITH JOINT SPARSITY PRIOR}
\label{sec:tddl_js}

 We now extend TDDL by using a joint sparsity (JS) prior (TDDL-JS). The joint sparsity prior \cite{Tropp_js, Cotter_js} enforces the sparse coefficients of the test pixel and its neighboring pixels within the neighborhood window to have row sparsity pattern, where all pixels are represented by the same atoms in the dictionary so that only few rows of the sparse coefficients matrix are nonzero.  The joint sparse recovery can be solved by the following Lasso problem

\begin{equation}
\label{eq:js}
\bA = \arg\min_{\bZ}{\lVert \bX -\bD\bZ \rVert_F^2 + \lambda \lVert \bZ \rVert_{1,2}  },
\end{equation}
where $\bA,\bZ\in \bbR^{N\times P}$ are sparse coefficient matrices and $\bX = \left[ \bx_1, \dots, \bx_P \right] \in \bbR^{M\times P}$ represents all the pixels within a neighborhood window centered on a test (center) pixel $\bx_c$. Define the label of the center pixel as $\by_c$. $P$ is the total number of pixels within the neighborhood window. $\lVert \bZ \rVert_{1,2} = \sum\limits_{i=1}^P \lVert \bZ_i \rVert_2$ is the $\ell_{1,2}$-norm of $\bZ$. $\bZ_i\in \bbR^{1\times P}$ is the $i^\text{th}$ row of $\bZ$. Many sparse recovery techniques are able to solve Eq. (\ref{eq:js}), such as the Alternating Direction Method of Multipliers \cite{admm}, Sparse Reconstruction by Separable Approximation (SpaRSA) \cite{sparsa} and Fast Iterative Shrinkage-Thresholding Algorithm (FISTA) \cite{Fista}.

Once the sparse code $\bA$ is obtained, the sparse codes $\balpha_c$ of the center pixel $\bx_c$ is projected on each of the $K$ decision planes of the classifier. The plane with the largest projection indicates the class that the center pixel $\bx_c$ belongs to, 
\begin{equation}
\label{eq:assign}
\text{identity}(\bx_c) = \arg\max_k \hat{\by}_k = \arg\max_k(\bW \balpha_c)_k,
\end{equation}
where $\balpha_c\in \bbR^{N}$ is the sparse coefficients of the center pixel.  In the training stage, it is expected that the projection of the decision plane corresponding to the class of the center pixel should be increased while other planes should be orthogonal to $\balpha_c$.   Therefore, given the training data $(\bX, \by_c)$, the classification error for the center pixel $\bx_c$ is defined as 
\begin{equation}
\label{eq:assign}
\calL(\by_c, \bW, \balpha_c(\bD, \bX)) = \lVert \by_c -  \bW \balpha_c\rVert_2^2 + \frac{\mu}{2}\lVert \bW \rVert_F^2,
\end{equation}
In order to update the dictionary $\bD$, we need to apply a chain rule similar to the one in Eq. (\ref{eq:chain_rule}):
\begin{equation}
\frac{\partial \calL}{\partial \bD} = \frac{\partial \calL}{\partial \bA}\frac{\partial \bA}{\partial \bD}.
\label{eq:chain_js}
\end{equation}
Now we focus on the difficult part $\frac{\partial \bA}{\partial \bD}$ of Eq. (\ref{eq:chain_js}). Employing the fixed point differentiation on Eq. (\ref{eq:js}), we have 
\begin{equation}
\frac{\partial   \lVert \bX - \bD\bA \rVert_F^2}{\partial \bA} \Big|_{\bA = \hat{\bA}} = - \lambda  \frac{\partial \lVert \bA \rVert_{1,2}}{\partial \bA} \Big|_{\bA = \hat{\bA}}.
\label{eq:diff_js_1}
\end{equation}
In the following part of this section, we omit the fixed point notation. Eq. (\ref{eq:diff_js_1}) is only differentiable when $\norm{\bA_i}_2 \neq \mathbf{0}$, where $\bA_i$ denotes the $i^\text{th}$ row of $\bA$.  At points where $\norm{\bA_i}_2 = \mathbf{0}$, the derivative  is not well defined, so we set  $\frac{\partial\norm{\bA_i}_2}{\partial \bA_i} = \mathbf{0}$. Denote $\tilde{\bA} = \bA_\Lambda \in \bbR^{N_{\Lambda}\times P}$, where $\Lambda$ is the active set such that  $\Lambda = \{ i : \norm{\bA_i}_2 \neq 0, i \in \{1,\dots, N\} \}$, $N_{\Lambda} = \lvert \Lambda \rvert$, $\bA_\Lambda$ is composed of active rows of $\bA$, and  $
\tilde{\bD}$ is the active atoms of $\bD$. Expanding the derivative of Eq. (\ref{eq:diff_js_1}) on both sides on the feasible points, 
\begin{equation}
\tbD^\top \left( \bX - \tbD \tbA   \right) = \lambda \left[ \frac{\tbA_1^\top}{\norm{\tbA_1}_2}, \dots, \frac{\tbA_{N_\Lambda}^\top}{\norm{\tbA_{N_\Lambda}}_2}  \right]^\top.
\label{eq:diff_js_2}
\end{equation}

\begin{algorithm}[!t]
\caption{Stochastic gradient descent algorithm for task-driven dictionary learning with joint sparsity prior}
%\footnotesize
\begin{algorithmic}[1]
%\caption{Task driven dictionary learning using group sparse recovery}
\fontsize{10}{10}
% \caption{Task driven dictionary learning with joint sparsity prior}

 %\SetAlgoLined
% input: Initial $\bD$ and $\bW$, training data set $(\by, \bx)$, initial step size $\rho_0$.

\Require Initial dictionary $\bD$ and classifier $\bW$. Parameter $\lambda$, $\rho$ and $t_0$.
 \For{$t = 1$ to $T$} 
   \State Draw one sample $(\bX, \by_c)$ from training set.
\State Find sparse sparse code $\bA$ according to Eq. (\ref{eq:js}).
\State Find the active set $\Lambda$ and define $N_{\Lambda} = \lvert \Lambda \rvert$
\begin{equation*}
\Lambda \leftarrow \{ i: \lVert \bA_i \rVert_2 \neq 0, i\in \{1,\dots, N\}\},
\end{equation*}
\NoNumber{where $\bA_i$ is the $i^{\text{th}}$ row of $\bA$.}
\State Compute $\bGamma \in \bbR^{N_{\Lambda}P\times N_{\Lambda}P}$
\begin{align*}
\bGamma &= \bGamma_1 \oplus \dots \oplus \bGamma_{N_\Lambda}, \\
\bGamma_i &=  \frac{\bI_P}{\lVert\tbA_i\rVert_2} - \frac{\tbA_i\tbA_i^\top}{\lVert\tbA_i\rVert_2^3},
i=1, \dots, N_\Lambda,
\end{align*}
\NoNumber where $\oplus$ is the direct sum of matrices. 
\State Compute $\bgamma\in \bbR^{N_{\Lambda}P}$
\begin{align*}
\bgamma &=  (\tbD^\top\tbD\otimes\bI_P + \lambda \bGamma)^{-\top}vec((\bW\hat{\bA}-\hat{\bY})^\top \tilde{\bW}).
\end{align*}
\NoNumber{where $vec(\cdot)$ and $\tilde{\bW}$ denote the vectorization operator and $\Lambda$ columns of $\bW$ respectively. }
\State Let $\bbeta \in \bbR^{N\times P}$. Set $\bbeta_{\Lambda^C}=\mathbf{0}$ and construct $\bbeta_{\Lambda}\in \bbR^{N_{\Lambda}\times P}$ that satisfies
\begin{equation*}
vec\left( \bbeta_{\Lambda}^\top \right) =\bgamma.
\end{equation*}

%\State Assign $\tilde{\bbeta}$ as the $c^\text{th}$ column of $\bbeta$.
\State Choose the learning rate $\rho_t \leftarrow \min(\rho, \rho \frac{t_0}{t} )$.
\State Update the parameters by gradient projection step
\begin{align*}
\bW &\leftarrow \bW - \rho_t\big((\bW\balpha_c -  \by)\balpha_c^\top + \mu \bW \big ), \\
\bD &\leftarrow \bD - \rho_t(-\bD \bbeta \bA^{\top} + (\bX - \bD\bA)\bbeta^\top),
\end{align*}
\NoNumber{and normalize every column of $\bD^{(t+1)}$ with respect to $\ell_2$-norm.}
\EndFor \\
\Return $\bD$ and $\bW$.

\end{algorithmic}
\label{algorithm:TDDL-JS} 
\end{algorithm}

Computing the derivative of Eq. (\ref{eq:diff_js_2}) with respect to $D_{mn}$ and transposing both sides
\begin{align} 
\frac{\partial \left\lbrace   \left( \bX - \bD\bA   \right)^\top \tbD \right\rbrace}{\partial D_{mn}}  =  \lambda \left[ \bGamma_{1} \frac{\partial \tbA_{1}^\top}{\partial D_{mn}}, \dots,\bGamma_{N_\Lambda} \frac{\partial \tbA_{N_\Lambda}^\top}{\partial D_{mn}} \right],
\label{eq:diff_js_3}
\end{align}
where $\bGamma_{i} = \frac{\bI_P}{\norm{\tbA_{i}}_2} - \frac{\tbA_{i}^\top \tbA_{i}}{\norm{\tbA_{i}}_2^3}$, $i = 1,\dots, N_\Lambda$. By vectorizing Eq. (\ref{eq:diff_js_3}), we have

\begin{align}
vec   \left( \frac{\partial \bX^\top \tbD}{\partial D_{mn}} -\tbA^\top\frac{\partial \tbD^\top \tbD}{\partial D_{mn}}  - \frac{\partial \tbA^\top}{\partial D_{mn}} \tbD^\top\tbD \right)
=  \lambda \cdot \bGamma vec\left( \frac{\partial \tbA^\top}{\partial D_{mn}} \right),
\label{eq:diff_js_4}
\end{align}
where $\bGamma = \bGamma_{1} \oplus \dots \oplus \bGamma_{N_\Lambda}$. From Eq. (\ref{eq:diff_js_4}), we reach the vectorization form of the derivative of $\tbA$ with respect to $D_{mn}$, given as
\begin{equation}
vec\left( \frac{\partial \tbA^\top}{\partial D_{mn}} \right) = \left( \tbD^\top \tbD \otimes \bI_P + \lambda \bGamma \right)^{-1} vec \left( \tbA^\top \frac{\partial \tbD^\top \tbD}{\partial D_{mn}}  + \frac{\partial  \bX^\top  \tbD}{\partial D_{mn}}\right).
\label{eq:diff_js_5}
\end{equation}

Now we can update the dictionary element-wise using Eq. (\ref{eq:diff_js_5}). In order to reach a more concise form for updating the dictionary, we perform algebraic transformations on Eq. (\ref{eq:chain_js}) and Eq. (\ref{eq:diff_js_5}), which are illustrated in Appendix \ref{append}. We illustrate the overall optimization for TDDL-JS in Algorithm \ref{algorithm:TDDL-JS}. It should be noted that in the Algorithm \ref{algorithm:TDDL-JS}, we define  $\hat{\bA} = [\mathbf{0}, \dots, \balpha_c, \dots, \mathbf{0}]\in \bbR^{N \times P}$ and $\hat{\bY} = [\mathbf{0}, \dots, \by, \dots, \mathbf{0}]\in \bbR^{K\times P}$. 

%\begin{align}
%\label{eq:g_projection}
%\bD &\leftarrow\bD - \rho_t \nabla_{\bD} \calL(\by, \bW, \bA(\bD, \bX)).
%\end{align}

%The dictionary is optimized by minimizing the classification error of the center pixel by using the stochastic gradient descent algorithm,
%\begin{align}
%\label{eq:g_projection}
%%\bW &\leftarrow \bW - \rho_t\nabla_{\bW} \calL(\by, \bW, \bA(\bD, \bX)) \\
%\bD &\leftarrow\bD - \rho_t\nabla_{\bD} \calL(\by, \bW, \balpha_c(\bD, \bX)).
%\end{align}

%We need to adopt the chain rule to compute $\nabla_{\bD} \calL$,
%\begin{align}
%\label{eq:chain_rule}
%\nabla_{\bD} \calL = \frac{\partial \calL}{\partial \balpha_c} \frac{\partial \balpha_c}{\partial \bD}.
%\end{align}
%The difficult part of computing the gradient $\nabla_{\bD} \calL$ is that $\bD$ is an implicit function of $\balpha_c$. Fortunately, it can be computed using fixed point equations \cite.

%In order to compute the $\nabla_{\bD}\calL$, the differentiability of $\calL$ needs to be investigated. Following in a similar manner as \cite{MairalTDDL,Yang}, the $\calL$ is found to have derivatives only when $\lVert \bA_i \rVert_2 \neq 0$, where $\bA_i$ denotes the $i^{\text{th}}$ row of $\bA$. Due to the space limitation, we omit any detailed derivation of the algorithm and directly illustrate the approach in Algorithm (\ref{algorithm:TDDL-JS}).

\section{TDDL WITH LAPLACIAN SPARSITY PRIOR}
\label{sec:laplacian}
The joint sparsity prior is a relatively stringent constraint on the sparse codes since it assumes that all the neighboring pixels have the same support as the center pixel.  The assumption of the joint sparsity prior can easily be  violated on non-homogeneous regions, such as a region that contains pixels from different classes. This makes choosing a proper neighborhood window size a difficult problem. When the window size is too large, the sparse codes of the non-homogeneous regions within the window are indiscriminative. On the other hand, the sparse codes are not stable if the window size is chosen to be too small. Ideally, we hope that the performance is  insensitive to both the choice of the window size and the topology of the image.  To achieve this requirement, we propose to enforce the  Laplacian sparsity (LP) prior (TDDL-LP) on the TDDL, where the degree of similarity between neighboring pixels can be utilized to push the sparse codes of the neighboring pixels that belong to the same class to be similar, instead of enforcing all the neighboring pixels to have a  similar sparse codes blindly.  The corresponding Lasso problem can be stated as follows

\begin{equation}
\label{eq:lap1}
\bA = \arg\min_{\bZ}{\lVert \bX -\bD\bZ \rVert_2^2 + \lambda \lVert \bZ \rVert_{1} + \gamma\sum^P\limits_{i,j}c_{ij}\lVert\bZ^i-\bZ^j\rVert^2_2},
\end{equation}
where $\bZ^i$ and $\bZ^j$ denote the $i^{\text{th}}$ and $j^{\text{th}}$ columns of $\bZ$. $c_{ij}$ is a weight whose value is proportional to the spectral similarity of $\bX^i$ and $\bX^j$, which are the $i^{\text{th}}$ and $j^{\text{th}}$ columns of $\bX$. $\gamma$ is a regularization parameter.

The Laplacian sparse recovery described by Eq. (\ref{eq:lap1})  in \cite{Gao} is able to discriminate pixels from different classes by defining an appropriate weighting matrix $\bC = [c_{ij}]\in \bbR^{P \times P}$. Additionally, it enforces both the support and the magnitude of sparse coefficients of similar spectral pixels to be similar, whereas the joint sparsity prior enforces sparse coefficients of all the  pixels within the neighborhood window to have the same support. Eq. (\ref{eq:lap1}) can be reformulated as

\begin{equation}
\label{eq:lap2}
\bA = \arg\min_{\bZ}{\lVert \bX -\bD\bZ \rVert_2^2 + \lambda \lVert \bZ \rVert_{1} + \gamma tr(\bZ\bL \bZ^{\top}) },
\end{equation}
where $\bL = \bB - \bC\in \bbR^{P\times P}$ is the Laplacian matrix \cite{Lux_laplacian}. $\bB = [b_{ij}]\in \bbR ^{P\times P}$  is a diagonal matrix such that  $b_{ii} = \sum\limits_{j} c_{ij} $. 

In this paper, we adopt the method of Sparse Reconstruction by Separable Approximation (SpaRSA) \cite{sparsa, Xiaoxia} to solve the Laplacian sparse coding problem.

\subsection{Sparse Recovery Algorithm}
 A modified feature sign search \cite{Gao} is capable of solving the optimization problem (\ref{eq:lap2}). It uses coordinate descent to update each column of $\bA$ iteratively. Although it gives plausible performance for the SRC-based HSI classification \cite{Xiaoxia}, it demands a high computational cost. The SpaRSA-based method can  achieve a similar optimal solution of Eq. (\ref{eq:lap2}) while being less computational burdensome. Despite the fact that our previous work \cite{Xiaoxia} has shown that the performance of the SRC-based approach for HSI classification can be largely influenced by the choice of specific optimization technique, we found that such influence is reasonably small when employing the dictionary-learning-based approach. Therefore, we use a SpaRSA-based method to solve the sparse recovery for the Laplacian sparsity prior. Although, SpaRSA is originally designed to solve the optimization of single-signal case, it can be easily extended to tackle the problem with multiple signals, such as the collaborative hierarchical Lasso (C-Hilasso) \cite{chilasso}. 

SpaRSA is able to solve optimization problems that have the following form
\begin{equation}
\min_{\bA\in\bbR^{N\times P}} f\left( \bA \right) + \lambda \psi \left( \bA \right),
\label{eq:sparsa_goal}
\end{equation}
where $f: \bbR^{N\times P} \rightarrow \bbR$ is a convex and smooth function,  $\psi: \bbR^{N\times P} \rightarrow \bbR$ is a separable regularizer and $\lambda$ is the regularization parameter. In the particular case of the Laplacian sparse recovery, the regularizer $\psi$ is chosen to be the $\ell_1-$norm, i.e. $\psi(\bA) = \norm{\bA}_1$, and the convex function $f$ is set as 
\begin{equation}
f\left( \bA \right) = \norm{\bX - \bD \bA}_F^2 + \gamma tr\left( \bA \bL \bA^\top \right).
\end{equation}
In order to search the optimal solution of Eq. (\ref{eq:sparsa_goal}), SpaRSA generates a sequence of iterations $ \bA^{\left( t \right)} $, $t = 1,2, \dots,$  by solving the following subproblem
\begin{equation}
\footnotesize
\bA^{(t+1)} \in \arg\min_{\bZ \in \bbR^{N\times P}} \left( \bZ - \bA^{(t)} \right)^\top \nabla f ( \bA^{(t)} ) + \frac{\eta^{(t)}}{2} \norm{\bZ - \bA^{(t)}}_F^2 + \gamma \psi \left( \bZ \right),
\label{eq:subproblem}
\end{equation}
where $\eta^{(t)}>0$ is a nonnegative scalar such that $\eta^{(t)} = \mu \eta^{(t-1)}$ and $\mu>1$. The Eq. (\ref{eq:subproblem}) can be simplified into the following form by eliminating the terms independent of $\bZ$
\begin{equation}
\min_{\bZ\in \bbR^{N\times P}} \frac{1}{2} \norm{\bZ - \bU^{(t)}}_F^2 + \frac{\gamma}{\eta^{(t)}} \psi (\bZ),
\label{eq:simplified_subproblem}
\end{equation}
where $\bU^{(t)} = \bA^{(t)} - \frac{1}{\eta^{(t)}} \nabla f (\bA^{(t)})$. The optimization problem in Eq. (\ref{eq:simplified_subproblem}) is separable element-wise, which can be reformulated into 
\begin{equation}
\min_{A_{ij}} \frac{1}{2} (z_{ij} - u_{ij}^{(t)})^2 + \frac{\lambda}{\eta^{(t)}} \psi_{ij} (\bZ), \forall i = 1,\dots, N \;\text{and} \;j = 1,\dots, P.
\label{eq:simplified_separate}
\end{equation}
The problem in Eq. (\ref{eq:simplified_separate}) has a unique solution and can be solved by the well-known soft thresholding operator $S(\cdot)$
\begin{equation}
z_{ij}^{*} = S_{\frac{\gamma}{\eta^{(t)}}}\left( u_{ij}^{(t)} \right) = sign(u_{ij}^{(t)})\max\{0, \lvert u_{ij} \rvert - \frac{\lambda}{\eta^{(t)}} \}.
\end{equation}

Comparing with the algorithm proposed in \cite{Gao}, which is based on the coordinate descent, Laplacian sparse recovery using  SpaRSA is more computationally efficient since it is able to cheaply search for a better descent direction $\nabla f(\bA)$.   The corresponding optimization is stated in Algorithm \ref{algorithm:LAP}.

\begin{algorithm}[!t]
\caption{Sparse recovery for Laplacian sparsity prior using SpaRSA}

\begin{algorithmic}[1]

\Require Dictionary $\bD$,  constants $\eta_0>0$, $0<\eta_{\min}<\eta_{\max}$, $\mu>1$
\State Set $t = 0$ and $\bA^{(0)} = \mathbf{0}$
 \Repeat 
\State choose $\eta^{(t)}\in [\eta_{\min}, \eta_{\max}]$
\State compute $\bU^{(t)} \leftarrow \bA^{(t)} - \frac{1}{\eta^{(t)}} \nabla f(\bA^{(t)})$.
\Repeat
\State $\bA^{(t)} \leftarrow S_{\frac{\gamma}{\eta^{(t)}}}\left( \bU^{(t)} \right), $
\State $\eta^{(t)} \leftarrow \mu\eta^{(t)}.$
\Until{stopping criterion is satisfied}
\State $t \leftarrow t+1.$
\Until{stopping criterion is satisfied} \\
\Return The optimal sparse coefficients $\bA^*$.

\end{algorithmic}
\label{algorithm:LAP}
\end{algorithm}

\subsection{Dictionary Update}
In order to adjust the dictionary, we now follow Eq. (\ref{eq:chain_js}) to derive $\frac{\partial \bA}{\partial \bD}$ using the fixed point differentiation. Applying differentiation on Eq. (\ref{eq:lap1}) on the fixed point $\hat{\bA}$
\begin{equation}
\frac{\partial   \lVert \bX - \bD\bA \rVert_F^2 + \gamma tr\left( \bA \bL \bA^\top \right)}{\partial \bA} \Big|_{\bA = \hat{\bA}} = - \lambda  \frac{\partial \lVert \bA \rVert_{1}}{\partial \bA} \Big|_{\bA = \hat{\bA}}.
\label{eq:diff_lp_1}
\end{equation}
In the following part, we omit the fixed point notation. By computing the derivation and then applying the vectorization on Eq. (\ref{eq:diff_lp_1}), we have
\begin{equation}
vec \left( \bD^\top \left( \bX - \bD \bA \right) - \gamma  \bA \bL \right) =  \lambda \cdot vec \left( sign \left( \bA \right) \right).
\label{eq:diff_lp_2}
\end{equation}
The differentiation $\frac{\partial vec \left( sign \left( \bA \right) \right)}{\partial D_{mn} }$ is not well defined on zero points of $vec \left( sign \left( \bA \right) \right)$. Similar as in TDDL-JS, we set the $i^\text{th}$ element $\frac{\partial vec \left( sign \left( \bA \right) \right)_i}{\partial D_{mn} } = 0$ when  $vec \left( sign \left( \bA \right) \right)_i = 0$. Denote the $\Lambda$ as the index set of nonzero elements of  $vec \left( sign \left( \bA \right) \right)$. Compute the derivative of Eq. (\ref{eq:diff_lp_2}) with respect to $D_{mn}$ 
\begin{equation}
\frac{ \partial \left\lbrace vec \left( \bD^\top \left( \bX - \bD \bA \right) - \gamma  \bA \bL \right)_\Lambda \right\rbrace }{\partial D_{mn}} = \mathbf{0},
\label{eq:diff_lp_3}
\end{equation}
which leads to
\begin{equation}
vec \left( \frac{\partial \bD^\top \bD}{\partial D_{mn}} \bA - \frac{\partial \bD^\top\bX}{\partial D_{mn}}  +  \bD^\top\bD \frac{\partial \bA}{\partial D_{mn}}  + \gamma \frac{\partial \bA}{\partial D_{mn}} \bL \right)_\Lambda = \mathbf{0}.
\label{eq:diff_lp_4}
\end{equation}
Now we reach the desired gradient
\begin{align}
&vec \left( \frac{\partial  \bA }{\partial D_{mn}} \right)_\Lambda = \nonumber \\
& \left(\bI_P \otimes \bD^\top\bD + \gamma \bL\otimes \bI_N \right)_{\Lambda, \Lambda}^{-1} vec \left( \frac{\partial \tbD^\top \tbD}{\partial D_{mn}} \tbA + \frac{\partial \tbD^\top \bX}{\partial D_{mn}}\right)_\Lambda.
\label{eq:diff_lp_5}
\end{align}
By applying algebraic simplification to Eq. (\ref{eq:diff_lp_5}), which is shown in Appendix \ref{append}, we reach the optimzation for TDDL-LP as stated in the Algorithm \ref{algorithm:TDDL-LAP}. It should be noted that $\hat{\bA}$ and $\hat{\bY}$ have the same definitions as those in Algorithm \ref{algorithm:TDDL-JS}.

\begin{algorithm}[!t]
\caption{Stochastic gradient descent algorithm for task-driven dictionary learning with Laplacian sparsity prior}
%\footnotesize
\begin{algorithmic}[1]
%\caption{Task driven dictionary learning using group sparse recovery}
\fontsize{10}{10}
% \caption{Task driven dictionary learning with joint sparsity prior}

 %\SetAlgoLined
% input: Initial $\bD$ and $\bW$, training data set $(\by, \bx)$, initial step size $\rho_0$.

\Require Initial dictionary $\bD$ and classifier $\bW$. Parameter $\lambda$, $\rho$ and $t_0$.
 \For{$t = 1$ to $T$} 
   \State Draw one sample $(\bX, \by_c)$ from training set.
\State Find  sparse code $\bA$ according to Eq. (\ref{eq:js}).
\State Find the active set $\Lambda$ 
\begin{equation*}
\Lambda \leftarrow \{ i:  vec(\bA)_i  \neq 0, i\in \{1,\dots, NP\}\},
\end{equation*}
\NoNumber{where $vec(\bA)_i$ is the $i^{\text{th}}$ element of $vec(\bA)$.}
\State Let $\bbeta \in \bbR^{N\times P}$. Set $vec(\bbeta)_{\Lambda^C}=\mathbf{0}$ and compute $vec(\bbeta)_{\Lambda}$ 
\begin{equation*}
vec(\bbeta)_{\Lambda} = (\bI_P \otimes \bD^\top \bD + \gamma\bL\otimes \bI_N)_{\Lambda,\Lambda}^{-1}vec(\bW^\top (\bW\hat{\bA}-\hat{\bY}) )_\Lambda,
\end{equation*}
\NoNumber{and $\otimes$ denotes the Kronecker product.}
\State Choose the learning rate $\rho_t \leftarrow \min(\rho, \rho \frac{t_0}{t} )$.
\State Update the parameters by gradient projection step
\begin{align*}
\bW &\leftarrow \bW - \rho_t\big((\bW\balpha_c -  \by)\balpha_c^\top + \mu \bW \big ), \\
\bD &\leftarrow \bD - \rho_t(-\bD\bbeta \bA^{\top} + (\bX - \bD\bA)\bbeta^\top),
\end{align*}
\NoNumber{and normalize every column of $\bD^{(t+1)}$ with respect to $\ell_2$-norm.}
\EndFor \\
\Return $\bD$ and $\bW$.

\end{algorithmic}
\label{algorithm:TDDL-LAP}
\end{algorithm}

\section{EXPERIMENTs}
\label{sec:results}
\subsection{Experiment Setup}
Cross-validation to obtain the optimal values for all parameters, including $\lambda, \epsilon, \gamma$ (sparse coding regularization parameters),  $\mu$ (regularization parameter for the classifier), $\rho_0$ (initial step size),  $N$ (dictionary size) and $P$ (number of neighboring pixels),  would introduce significant computational cost. Instead, we search for the optimal values for the above parameters according to the following procedure.

\begin{itemize}
\item The candidate dictionary sizes are from $5$ to $10$ atoms per class. The choice of dictionary size depends on the classification performance and computational cost. In our experiment, we set the dictionary size to be $5$ atoms per class.

\item Searching for the optimal window size and the regularization parameters would be cumbersome. Empirically, we found that the optimal regularization parameters are less likely to be affected by the choice of the window size.  Therefore, for each image, we fix the window size to be $3\times 3$ in order to save computational resource during the search of the optimal regularization parameters. Candidate regularization parameters are $\left\lbrace 10^{-3},10^{-2},10^{-1} \right\rbrace$. 

\item The possible candidate window sizes are $3\times 3$,  $5\times 5$, $7\times 7$ and $9\times 9$. We search for the optimal window size for each image after finding the optimal regularization parameters.

\end{itemize}

\begin{table}[ht]
        \caption{Parameters Used in the Paper  }
        \centering
\begin{tabular}[ht]{ c | c | c | c   }
\hline
Structured Priors  &$\lambda$  &$\gamma$  &$\rho$  \\
\hline
\textbf{$\ell_1$}  &$10^{-2}$ & \backslashbox &$10^{-2}$ \\
\hline
JS  &$10^{-2}$ &\backslashbox  &$10^{-3}$ \\
\hline
LP   &$10^{-2}$ &$10^{-3}$  &$ 10^{-1}$ \\
\hline
\end{tabular}
\label{table:indian_parameter}
\end{table}

 Computing the gradient for a single training sample at each iteration of Algorithm \ref{algorithm:TDDL-JS} or \ref{algorithm:TDDL-LAP} will make the algorithm converge very slowly. Therefore, following the previous work \cite{Mairal,MairalTDDL}, we implement the two proposed algorithms with the mini-batch method, where the gradients of multiple training samples are computed in each iteration.   For the unsupervised learning methods, the batch size is set to $200$. For the supervised learning methods, the batch size is set to $100$  and $t_0=T/10$. We search the optimal regularization parameters for each image and found that their optimal values are coincidentally the same.  The reason could be due to our choice of a large interval for the search grid.  The regularization parameters used in our paper are shown in Table \ref{table:indian_parameter}. We set $\mu=10^{-4}$. As a standard procedure, we evaluate the classification performance on HSI image using the overall accuracy (OA), average accuracy (AA) and kappa coefficient ($\kappa$). The classification methods that are tested and compared are SVM, SRC, SRC with joint sparsity prior (SRC-JS), SRC with Laplacian sparsity prior (SRC-LP), unsupervised dictionary learning (ODL), unsupervised dictionary learning with joint or Laplacian sparsity prior (ODL-JS, ODL-LP), TDDL, TDDL-JS and TDDL-LP.
During the testing stage, all training pixels are excluded from the HSI image, which means there may be some `holes' (training pixels deleted) inside a neighborhood window. This is reasonable since we do not want the classification results to be affected by the spatial distribution of the labelled samples. We use SPAMS toolbox \cite{spams} to perform the  joint sparse recovery via the Fast Iterative Shrinkage-Thresholding Algorithm \cite{Fista}. The sparse recovery for SRC-based methods are performed via the Alternating Direction Method of Multipliers \cite{admm}.  The modified SpaRSA shown in Algorithm \ref{algorithm:LAP} is used to solve the Laplacian sparse recovery problem. 

For the unsupervised dictionary learning methods, the dictionary is initialized by randomly choosing a subset of the training pixels from each class and updated  using the online dictionary learning (ODL) procedure in \cite{Mairal}. The classifier's parameter are then obtained by using a multi-class linear regression.   For the supervised dictionary learning methods, the dictionary and classifier's parameter are initialized by the training results of ODL for the unsupervised method.

\begin{figure}[ht]
\centering

\begin{minipage}[b]{0.45\linewidth}
\centering
\includegraphics[width=\textwidth]{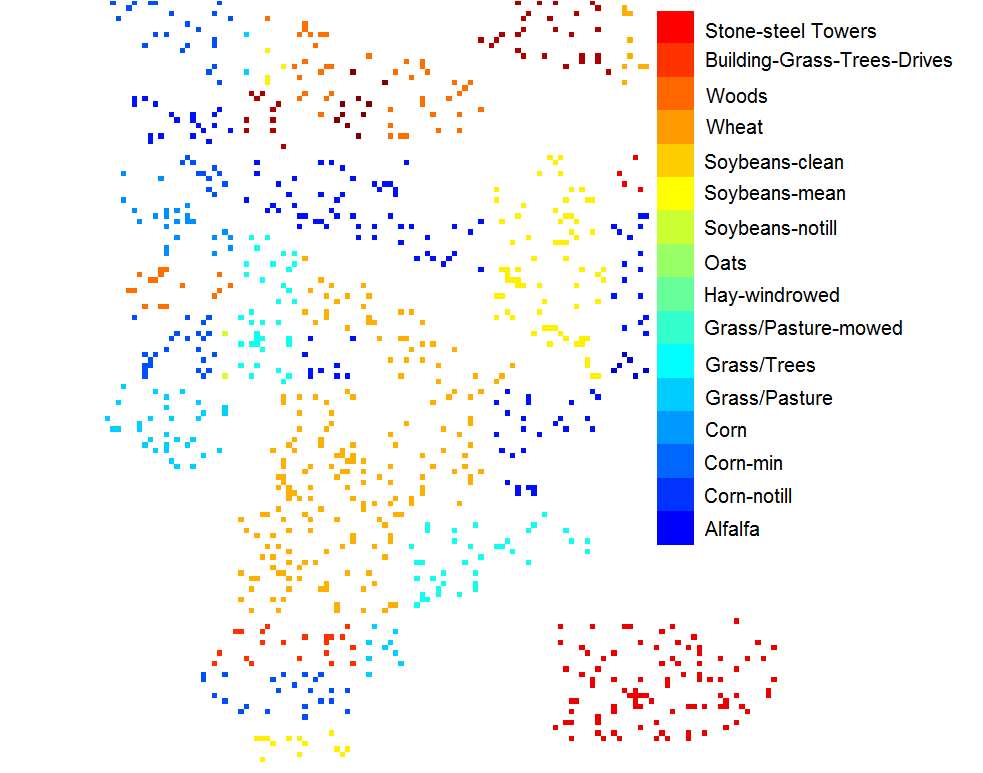}
\caption*{(a)}
\end{minipage}
\hspace{0.3cm}
\begin{minipage}[b]{0.45\linewidth}
\centering
\includegraphics[width=\textwidth]{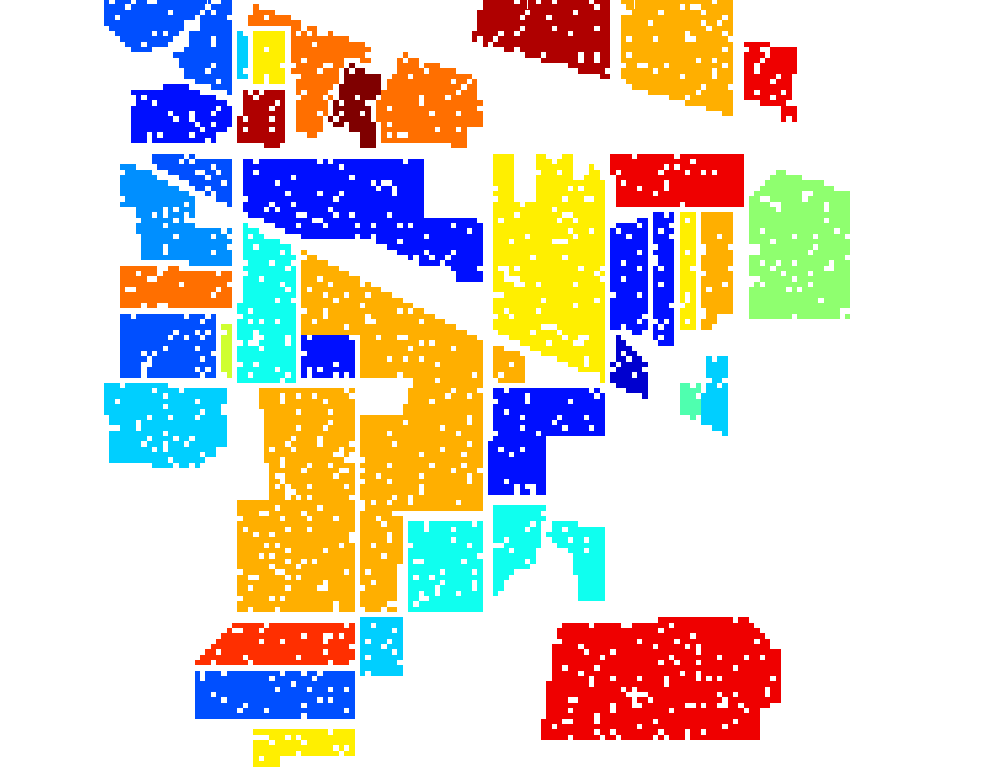}
\caption*{(b)}
\end{minipage}

%\captionsetup{justification=left}
%\captionsetup{justification=raggedright, singlelinecheck=false}
\caption{(a) Training sets and (b) test sets of the Indian Pine image.}
\label{fig:indian_pine_samples}
\end{figure}

\begin{table}[ht]

      \caption{Number of training and test samples for the Indian Pine image}
      \centering    
        \begin{tabular}[ht]{ c| c || c | c  }
\hline
Class \#  &Name & Train & Test \\
\hline
 1    &Alfalfa  &6 & 48 \\
 2  &Corn-notill &137 &1297 \\
 3  &Corn-min &80& 754   \\
 4   &Corn &23 &211 \\
 5  &Grass/Pasture  &48&449 \\
 6  &Grass/Trees &72 & 675\\
 7  &Grass/Pasture-mowed &3 & 23 \\
 8  &Hay-windrowed  &47 &442\\
 9  &Oats &2& 18 \\
10 &Soybeans-notill &93 & 875 \\
11 &Soybeans-min  &235 & 2233 \\
12 &Soybean-clean   &59 &555\\
13  &Wheat &21 &191 \\
14 &Woods  &124&1170\\
15 &Building-Grass-Trees-Drives &37 &343 \\
16  &Stone-steel Towers &10& 85 \\
\hline
\multicolumn{2}{c||}{Total}  &997 &9369\\
\hline

\end{tabular}
\label{table:indian_pine_sample}
\end{table}

\begin{figure}
\centering
\includegraphics[width=0.6\linewidth]{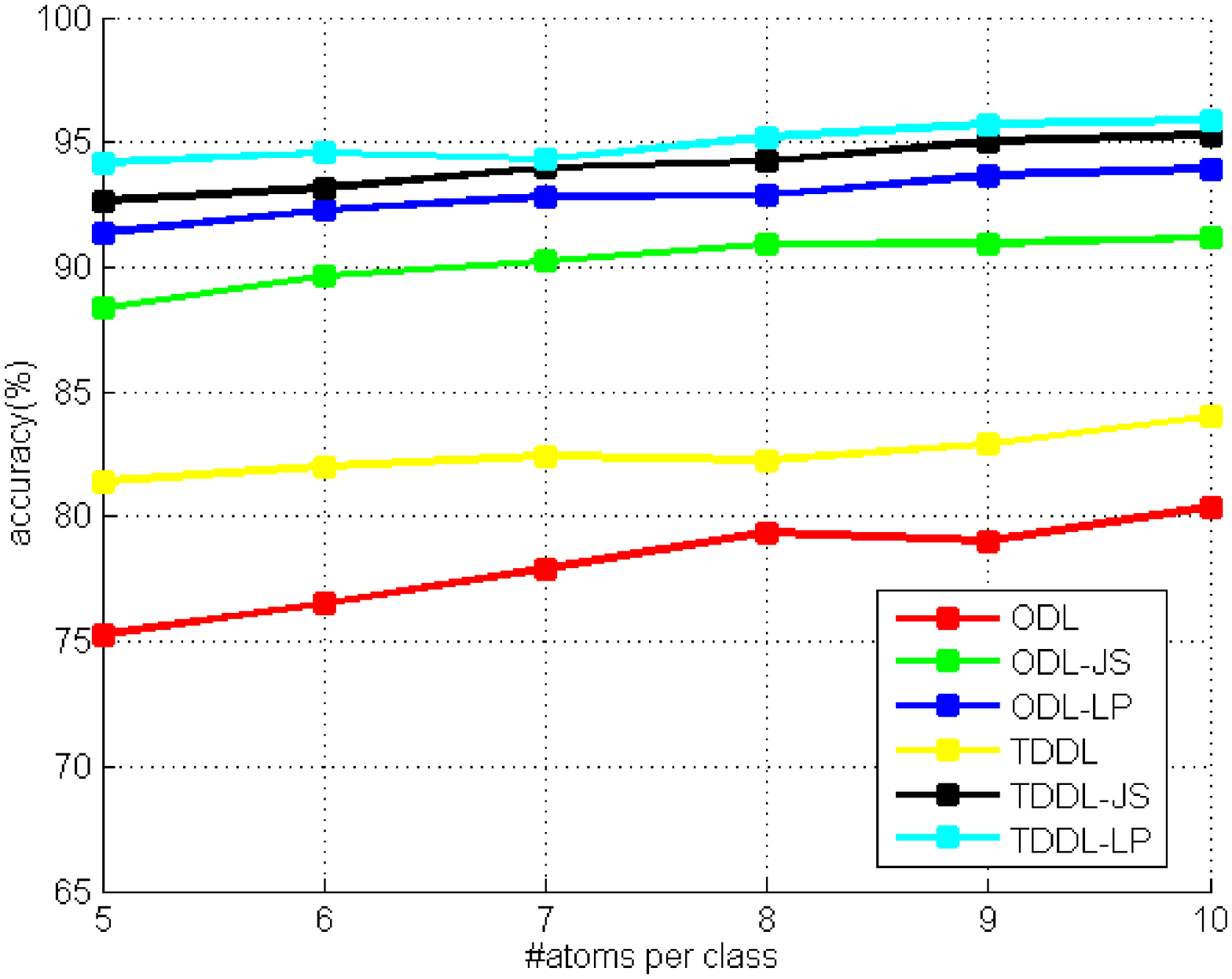}
\caption{The result with different dictionary sizes for the Indian Pine image. }
\label{fig:dictionarysize}
\end{figure}

\begin{figure}
\centering
\includegraphics[width=0.6\linewidth]{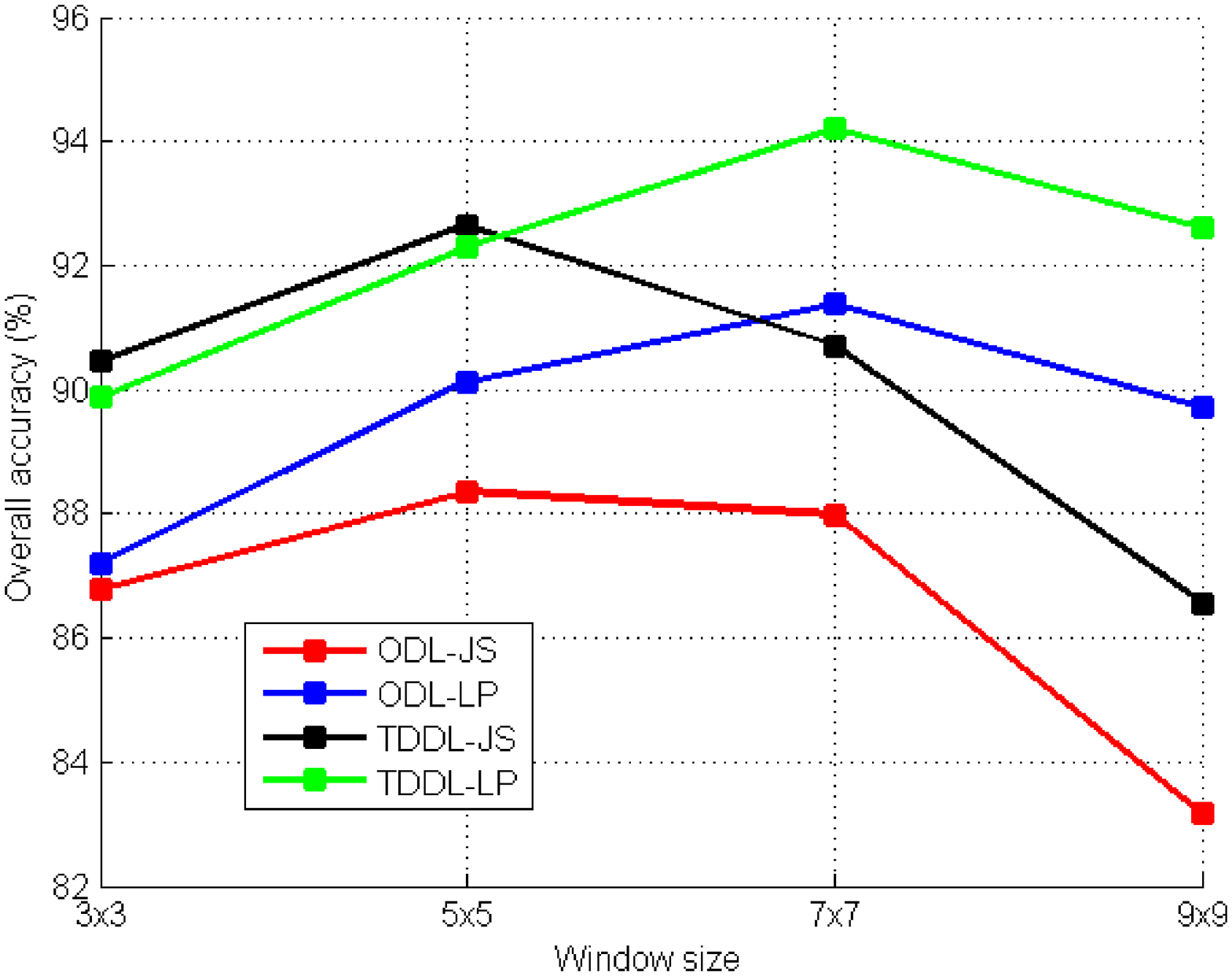}
\caption{The effect of different window sizes for the Indian Pine image. The dictionary size is fixed at five atoms per class.}
\label{fig:windowsize}
\end{figure}

\begin{figure*}[ht]
\centering

\begin{minipage}[b]{0.18\linewidth}
\centering
\includegraphics[width=\textwidth]{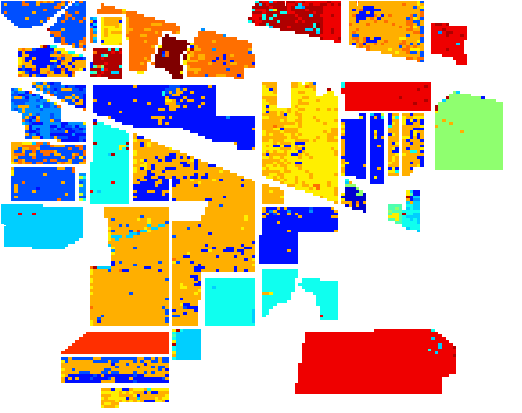}
\caption*{\tiny (a) SVM, OA = $64.94\%$}
\end{minipage}
\hspace{0.1cm}
\begin{minipage}[b]{0.18\linewidth}
\centering
\includegraphics[width=\textwidth]{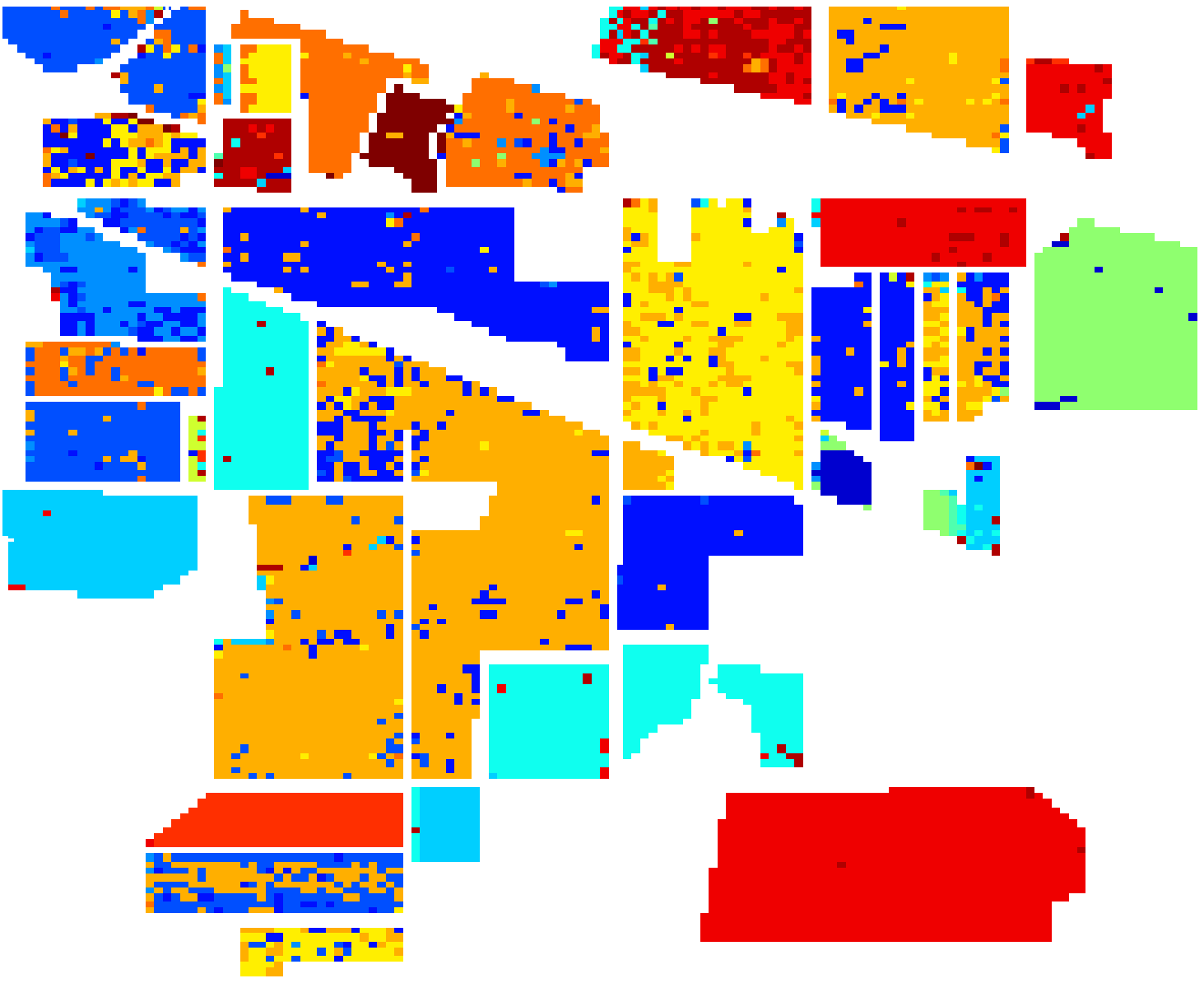}  %SRC
\caption*{\tiny (b) SRC, OA = $71.17\%$}
\end{minipage}
\begin{minipage}[b]{0.18\linewidth}
\centering
\includegraphics[width=\textwidth]{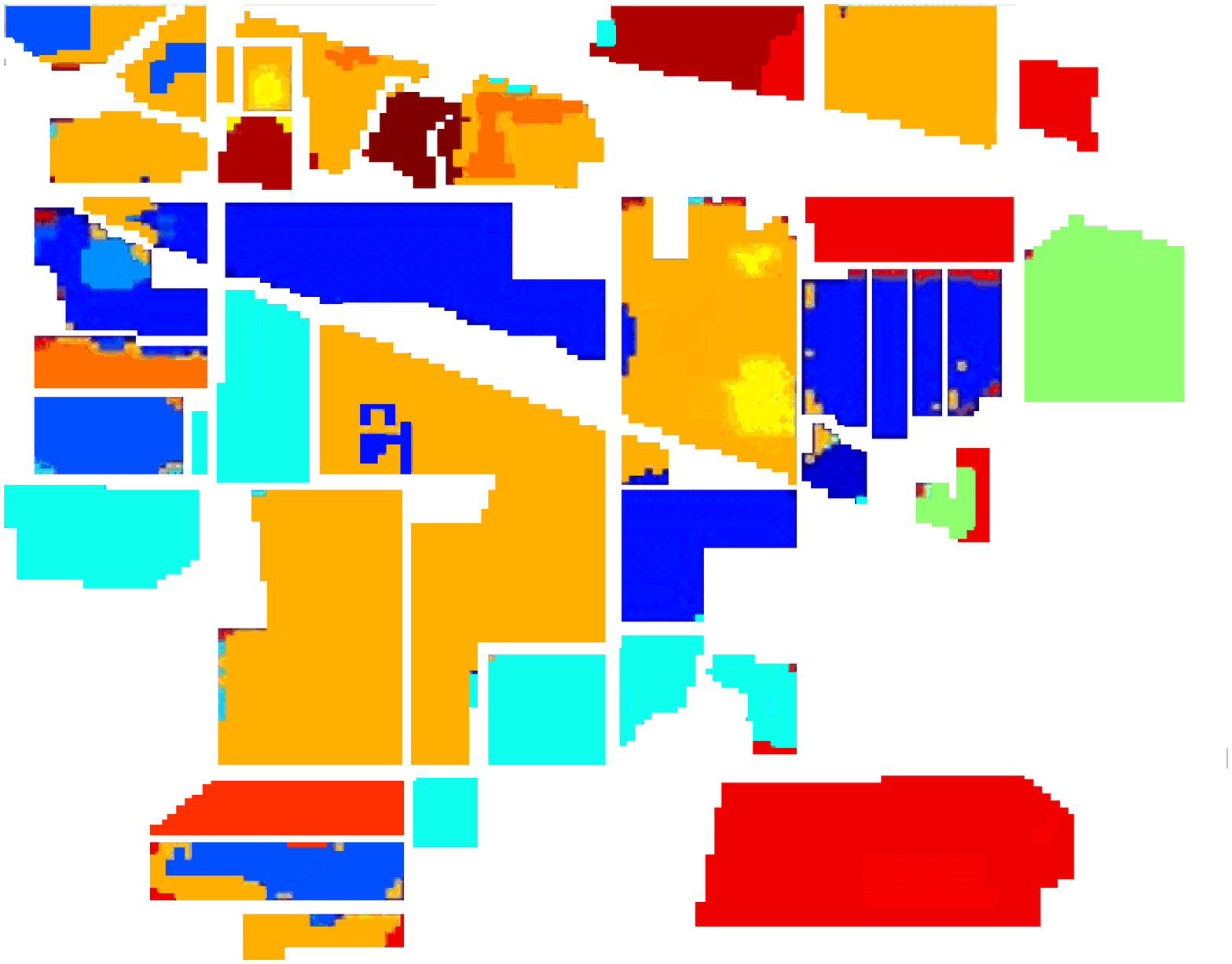}  %SRC-JS
\caption*{\tiny (c) SRC-JS, OA = $76.41\%$}
\end{minipage}
\begin{minipage}[b]{0.18\linewidth}
\centering
\includegraphics[width=\textwidth]{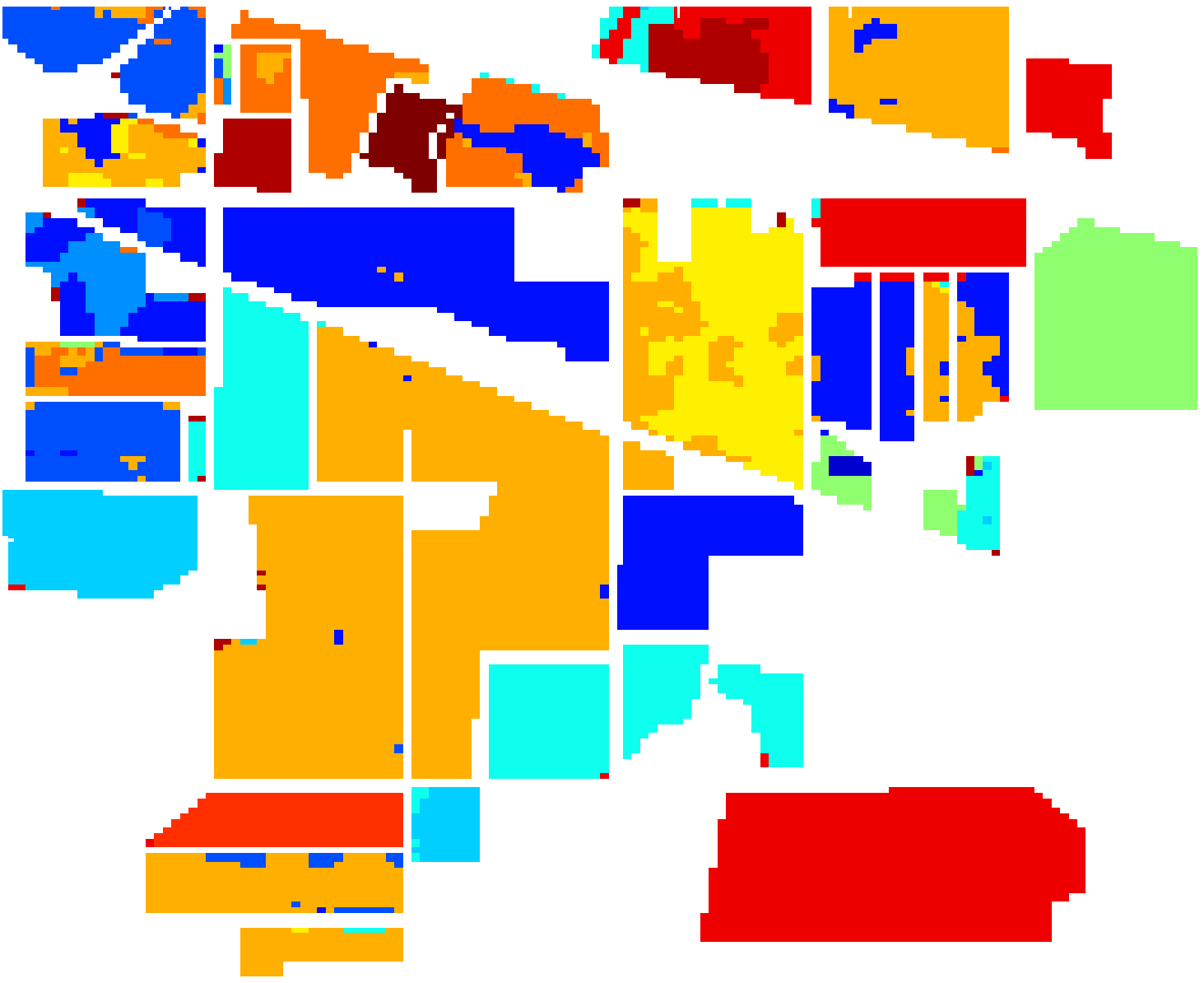}   %SRC-LP
\caption*{\tiny (d) SRC-LP, OA = $79.40\%$}
\end{minipage}
\hspace{0.1cm}
\begin{minipage}[b]{0.18\linewidth}
\centering
\includegraphics[width=\textwidth]{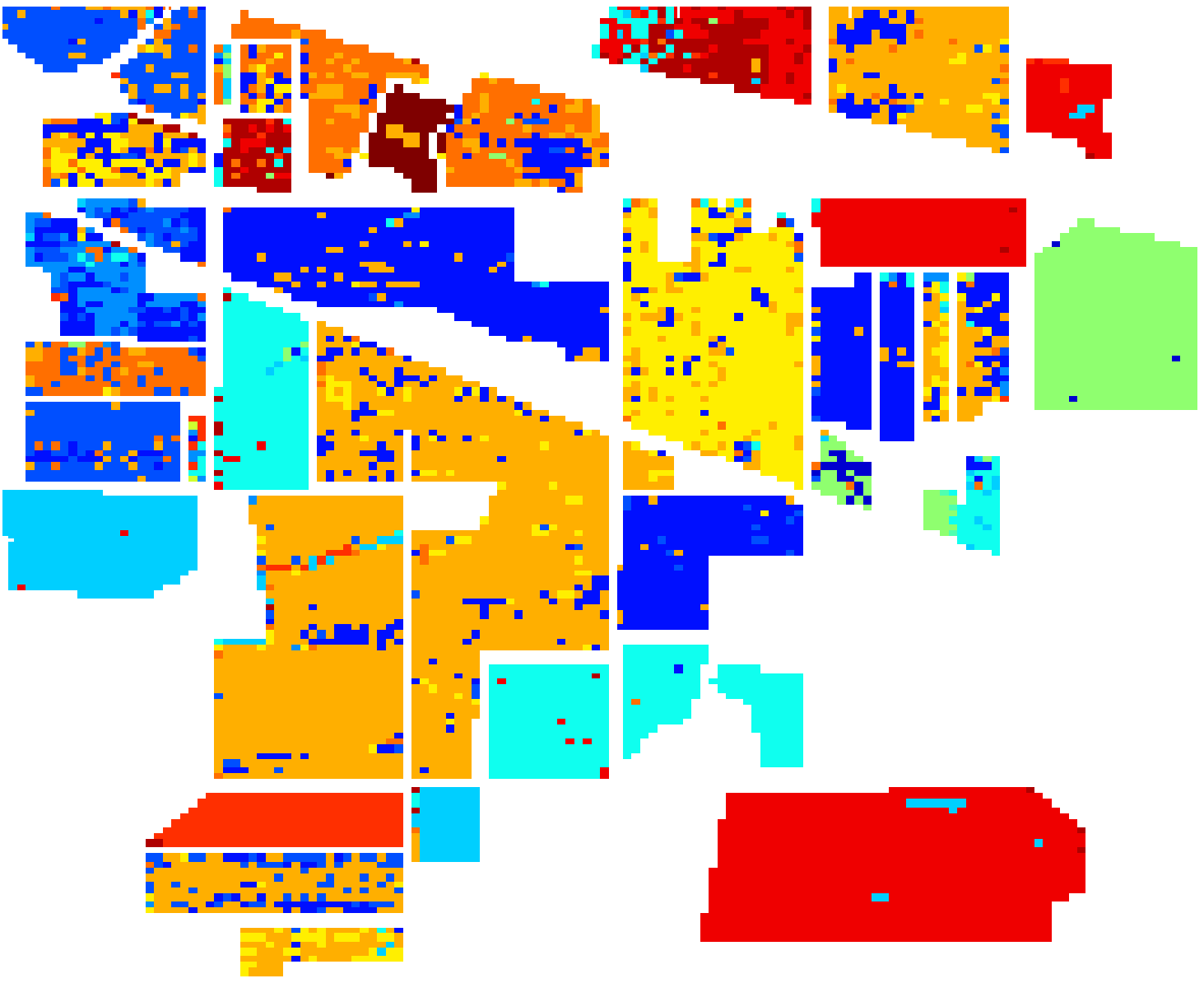}   %l1-unsup
\caption*{\tiny (e) ODL, OA = $71.04\%$}
\end{minipage}

\centering
\begin{minipage}[b]{0.18\linewidth}
\centering
\includegraphics[width=\textwidth]{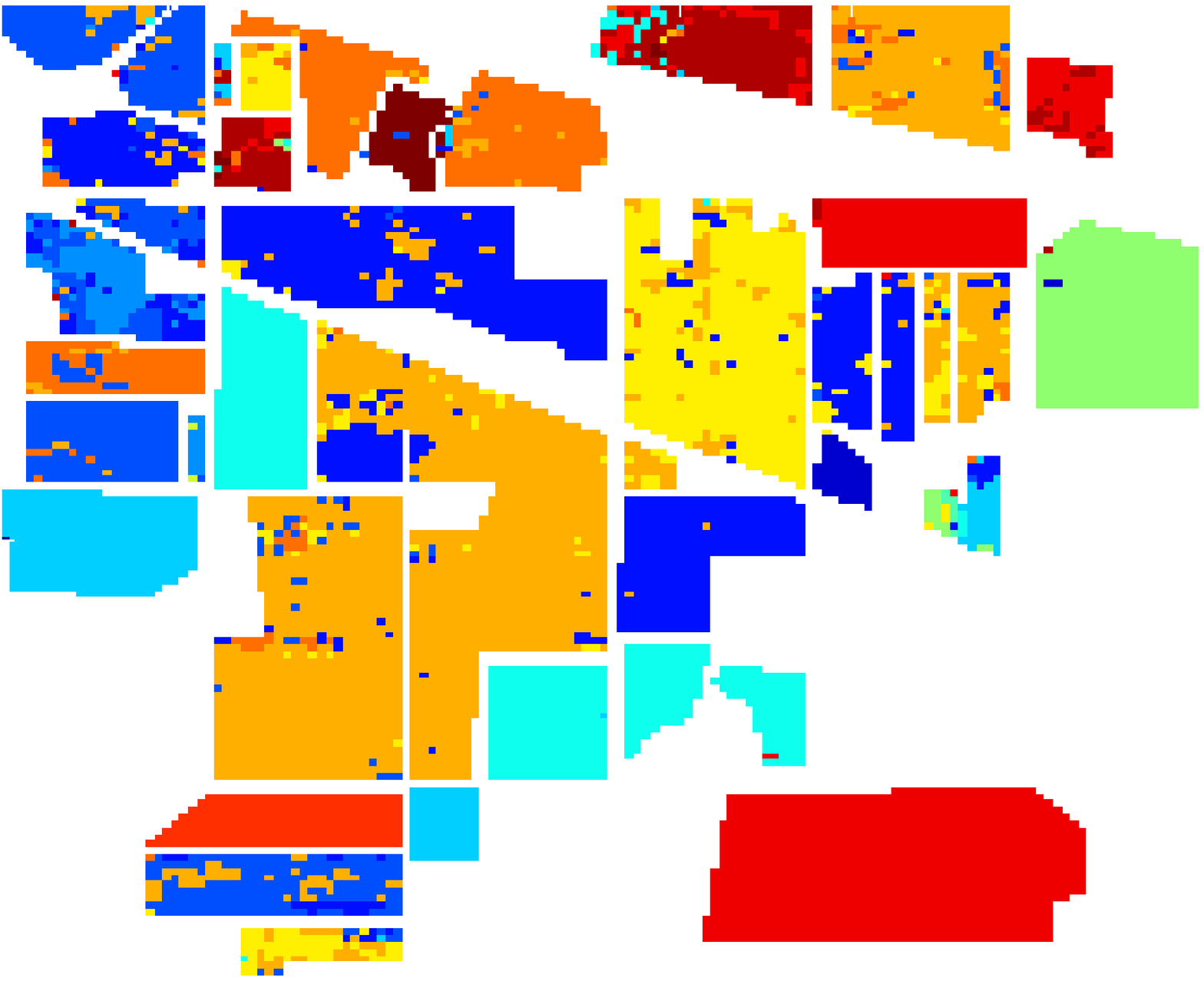}  %for ODL-JS
\caption*{\tiny (f) ODL-JS, OA = $88.36\%$}
\end{minipage}
\hspace{0.1cm}
\begin{minipage}[b]{0.18\linewidth}
\centering
\includegraphics[width=\textwidth]{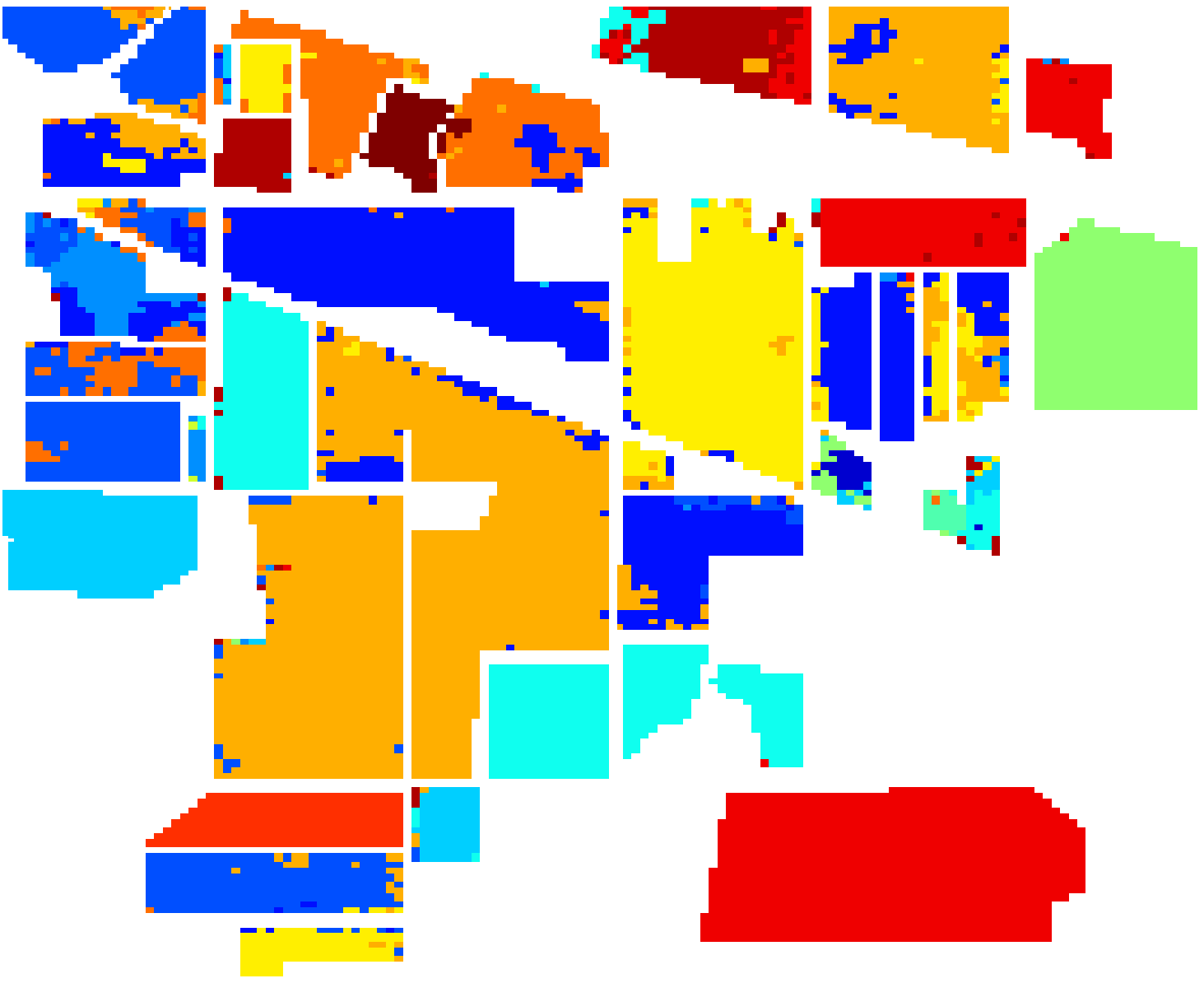}
\caption*{\tiny (g) ODL-LP, OA = $91.39\%$}
\end{minipage}
\hspace{0.2cm}
\begin{minipage}[b]{0.18\linewidth}
\centering
\includegraphics[width=\textwidth]{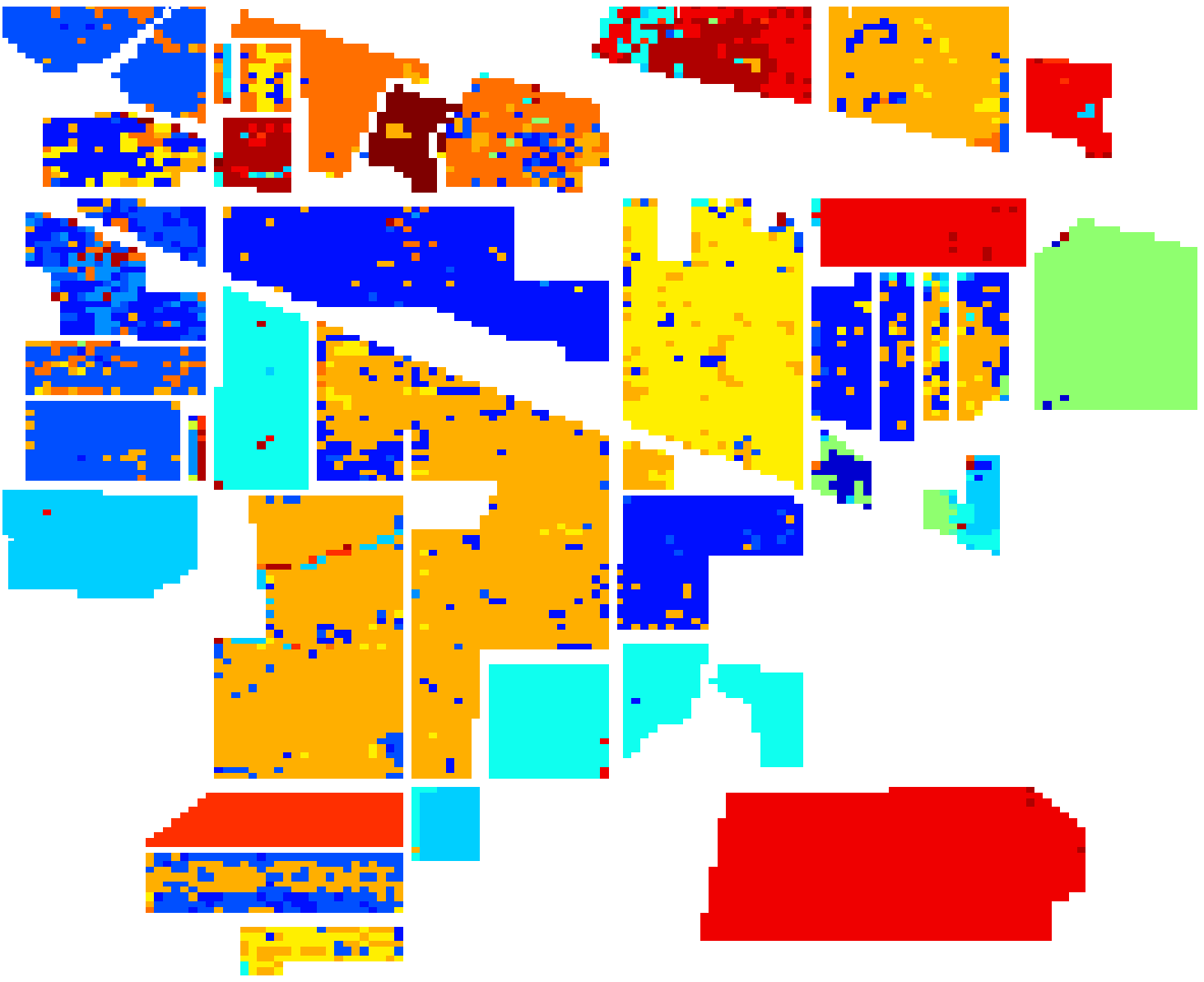}   %l1-sup
\caption*{\tiny (h) TDDL, OA = $81.43\%$}
\end{minipage}
\hspace{0.2cm}
\begin{minipage}[b]{0.18\linewidth}
\centering
\includegraphics[width=\textwidth]{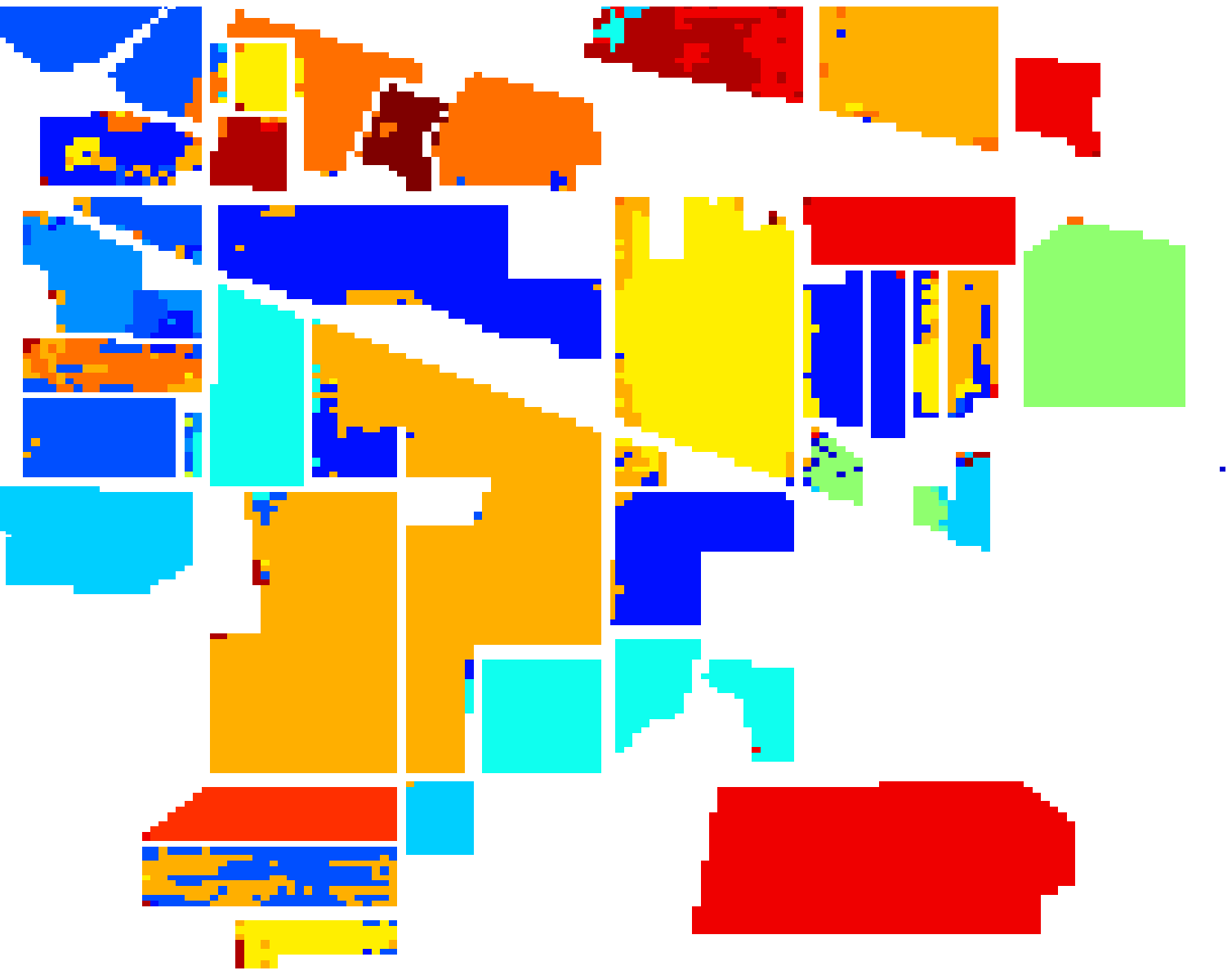} %js-tddl
\caption*{\tiny (i) TDDL-JS, OA = $92.65\%$}
\end{minipage}
\begin{minipage}[b]{0.18\linewidth}
\centering
\includegraphics[width=\textwidth]{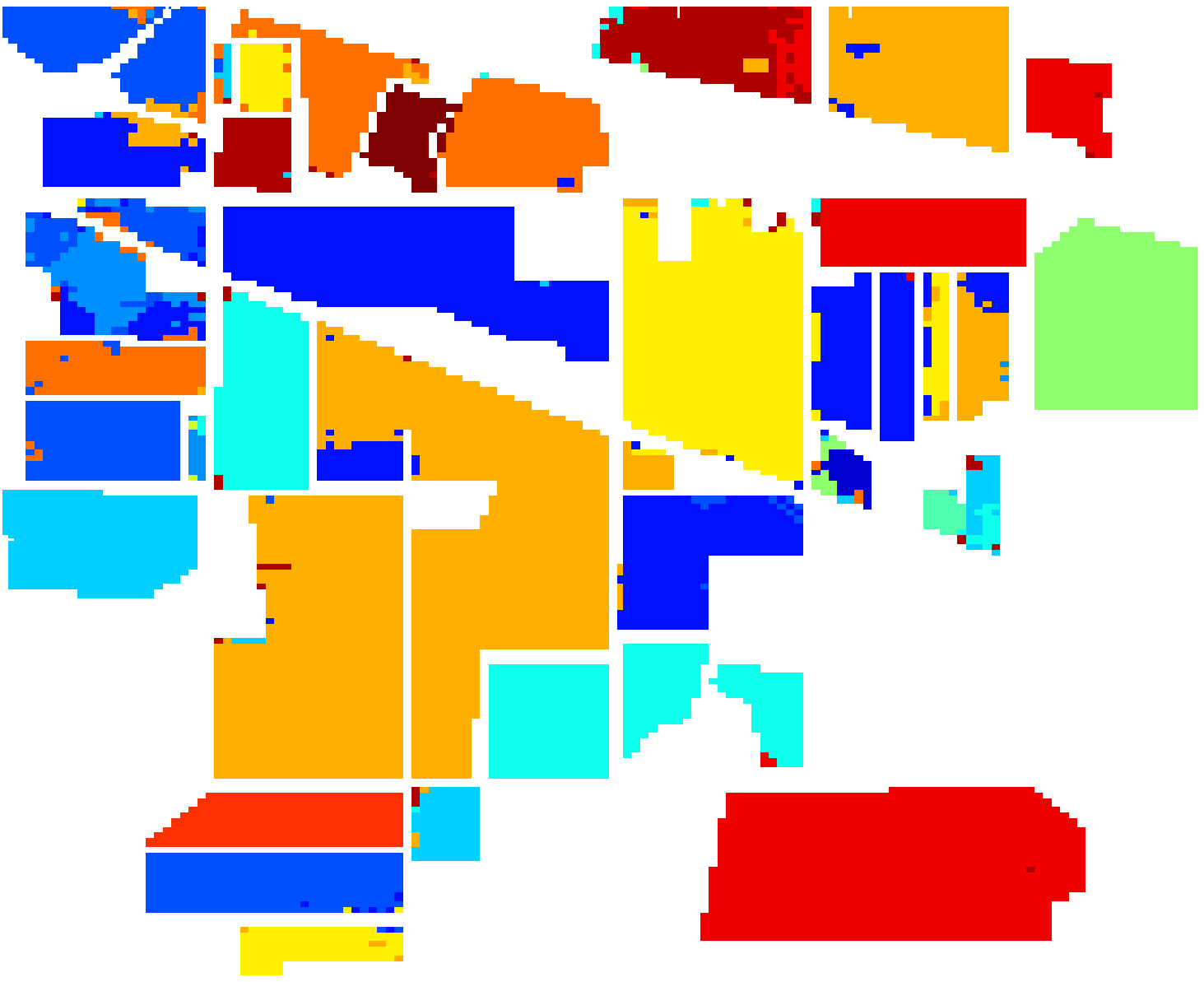}
\caption*{\tiny (j) TDDL-LP, OA = $94.20\%$}
\end{minipage}
\hspace{0.2cm}

%\captionsetup{justification=left}
\captionsetup{justification=raggedright, singlelinecheck=false}
\caption{Classification map of the Indian Pine image obtained by (a) SVM, (b) SRC, (c) SRC-JS, (d) SRC-LP, (e) ODL, (f) ODL-JS, (g) ODL-LP, (h) TDDL,  (i) TDDL-JS and (j) TDDL-LP.}
\label{fig:indian_pine}
\end{figure*}

\begin{table*}

      \centering
        \caption{Classification accuracy ($\%$) for the Indian Pine image}
        
\begin{tabular}[ht]{  c || c |  c  c  c | c   c   c   c   c   c   }

\hline
\multicolumn{2}{c|}{Dictionary Size} &\multicolumn{3}{c|}{$N=997$} &\multicolumn{6}{c}{$N=80$}\\
\hline
Class    & SVM &SRC  &SRC-JS &SRC-LP & ODL & ODL-JS  &ODL-LP & TDDL & TDDL-JS  &TDDL-LP \\
\hline
 1  &77.08  &68.75   &79.17 &82.42  &75.00  &\textbf{97.92} &70.83 &50.00  &35.42 &56.25 \\
 2  &84.96  &58.84   &81.94 &81.34  &59.69  &91.24 &94.26 &84.03  &\textbf{94.57} &93.95\\
 3  &62.67  &24.40   &56.67 &47.35   &62.93  &81.20 &84.40 &69.73  &84.13          &\textbf{92.13}\\
 4  &8.57   &49.52   &27.62 &49.76  &23.81  &47.62 &61.90 &14.76  &\textbf{79.05} &46.19\\
 5  &77.18  &81.88   &85.46 &83.96  &82.55  &\textbf{93.29} &92.62 &89.04  &90.16 &90.83\\
 6  &91.82  &96.88   &98.36 &97.48   &88.24  &99.55 &98.96 &98.66  &\textbf{99.55} &98.96\\
 7  &13.04  &0.00    &0.00 &0.00   &4.35   &\textbf{17.39} &0.00 &0.00   &0.00 &\textbf{95.65}\\
 8  &96.59  &96.59   &\textbf{100.00} &99.55  &96.36  &99.32 &99.32 &99.09  &100.00 &100.00\\
 9  &0.00   &\textbf{5.56}    &0.00 &0.00   &0.00   &0.00  &0.00 &0.00  &0.00 &0.00\\
10  &71.30  &24.00   &18.94 &31.89  &67.51  &77.73 &91.04 &72.90  &90.13 &\textbf{94.03}\\
11  &35.25  &96.22   &91.63 &94.58  &67.94  &88.25 &94.10 &85.46  &96.22 &\textbf{97.37}\\
12  &42.39  &32.97   &45.29  &64.68  &80.62  &88.59  &83.15 &59.06  &86.78 &\textbf{95.47}\\
13  &91.05  &98.95   &99.47  &99.48  &95.79  &\textbf{100.00}  &\textbf{100.00}  &\textbf{100.00} &\textbf{100.00} &\textbf{100.00}\\
14  &94.85  &98.97   &98.97 &99.49  &87.20  &97.77 &99.14 &98.11  &\textbf{99.40} &\textbf{99.40}\\
15  &30.70  &49.71   &55.85 &63.84  &32.16  &70.76  &67.84 &47.66  &77.78 &\textbf{82.75}\\
16  &27.06  &88.24   &95.29 &97.65  &69.41  &96.47 &85.88 &92.94  &91.76 &\textbf{98.82}\\
\hline
\textbf{OA}[$\%$]  & 64.94  &71.17    & 76.41 &79.40 &71.04 &88.36 &91.39 &81.43  &92.65 & \textbf{94.20} \\
\textbf{AA}[$\%$]  & 56.53  &60.72    & 64.67 &64.67 &62.10 &77.94 &82.18 &66.43 &76.56 &\textbf{83.86}\\
$\mathbf{\kappa}$  & 0.647  &0.695  & 0.737 &0.712 &0.691 &0.851 &0.907 &0.8087 &0.924 &\textbf{0.940}\\
\hline

\end{tabular}
\label{table:indian_pine_result}
\end{table*}

\subsection{Classification on AVIRIS Indian Pine Dataset}
We first perform HSI classification on the Indian Pine image, which is generated by Airborne Visible/Infrared Imaging Spectrometer (AVIRIS). Every pixel of the Indian Pine consists of 220 bands ranging from $0.2$ to $2.4 \mu$m, of which 20 water absorption bands are removed before classification. The spatial dimension of this image is $145\times 145$.  The image contains 16 ground-truth classes, most of which are crops, as shown in Table \ref{table:indian_pine_sample}. We randomly choose 997 pixels ($10.64\%$ of all the interested pixels)  as the training set and the rest of the interested pixels for testing.

The total iterations of unsupervised and supervised dictionary learning methods are  set to $15$ and $200$ respectively for this image. The classification results with varying dictionary size $N$ are shown in Fig. \ref{fig:dictionarysize}. In most cases, the classification performance increases with  the increment in  the dictionary size. All methods attain their highest OA when the dictionary size is $10$ atoms per class. The OA of ODL-JS, ODL-LP, TDDL-JS and TDDL-LP do not change much when the dictionary size increase from $5$ to $10$ atoms per class. Therefore, it is reasonable to set the dictionary size to be $5$ atoms per class by taking computational cost into account. Fig. \ref{fig:dictionarysize} also suggests that a plausible performance can be obtained even when the dictionary is very small and not over-complete. The classification performance with respect to the window size is demonstrated in Fig. \ref{fig:windowsize}. Using a window size of $5\times 5$, ODL-JS and TDDL-JS achieves the highest OA of $88.36\%$ and $92.65\%$, respectively. When the window size is set to $7\times 7$, the ODL-LP and TDDL-LP reach their highest OA $=91.39\%$ and OA $=94.20\%$, respectively. ODL-JS and TDDL-JS reach better performance when the window size is not larger than $5\times 5$. The TDDL-LP outperforms all other methods when the window size is $7\times 7$ or larger. Since a larger window size has more chances to include non-homogeneous regions, it verifies our argument that the Laplacian sparsity prior works better for classifying pixels lying in the non-homogeneous regions. 

Detailed classification results of various methods are shown in Table \ref{table:indian_pine_result} and visually displayed in Fig. \ref{fig:indian_pine}.   The OA of ODL-LP reaches $91.39\%$, which is more than $20\%$ higher than that of ODL and $3\%$ higher than that of ODL-JS.  The TDDL-LP has the highest classification accuracy for most classes. Most methods have $0\%$ accuracy for class $9$ since there are too few training samples in this class. The overall performance of TDDL-JS and TDDL-LP have at least $13\%$ improvement over the other conventional dictionary learning techniques. TDDL-LP significantly outperforms other methods on the classes that occupy small regions in the image. The class $7$ (Grass/Pasture-mowed), lying in a non-homogeneous region, has only $3$ training samples and $23$ test samples. The TDDL-LP is capable of correctly classify $95.65\%$ test samples while the second highest accuracy is only $17.39\%$. We notice that the AA of both ODL-LP ($82.18\%$) and TDDL-LP ($83.86\%$) are at least $4\%$ higher than that of the other methods. This also suggests that the Laplacian-sparsity-enforced dictionary learning methods work better on non-homogeneous regions, since the AA can only attain high value when both the most regions reach high accuracy.

%\begin{figure}
%\centering
%\includegraphics[width=0.4\linewidth]{white.eps}
%\caption*{The effect of using structured sparsity priors on convergence of the dictionary learning.(FFS, Fista, FFS-LP, Fista-JS)}
%\end{figure}

\begin{figure}[ht]
\centering

\begin{minipage}[b]{0.35\linewidth}
\centering
\includegraphics[width=\textwidth]{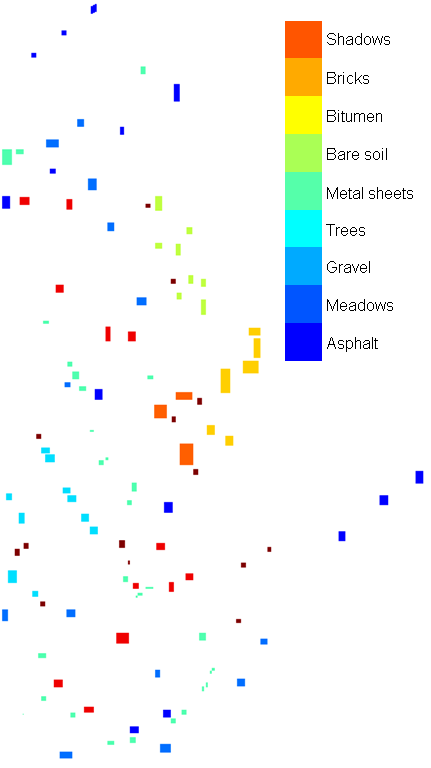}
\caption*{(a)}
\end{minipage}
\hspace{0.3cm}
\begin{minipage}[b]{0.35\linewidth}
\centering
\includegraphics[width=\textwidth]{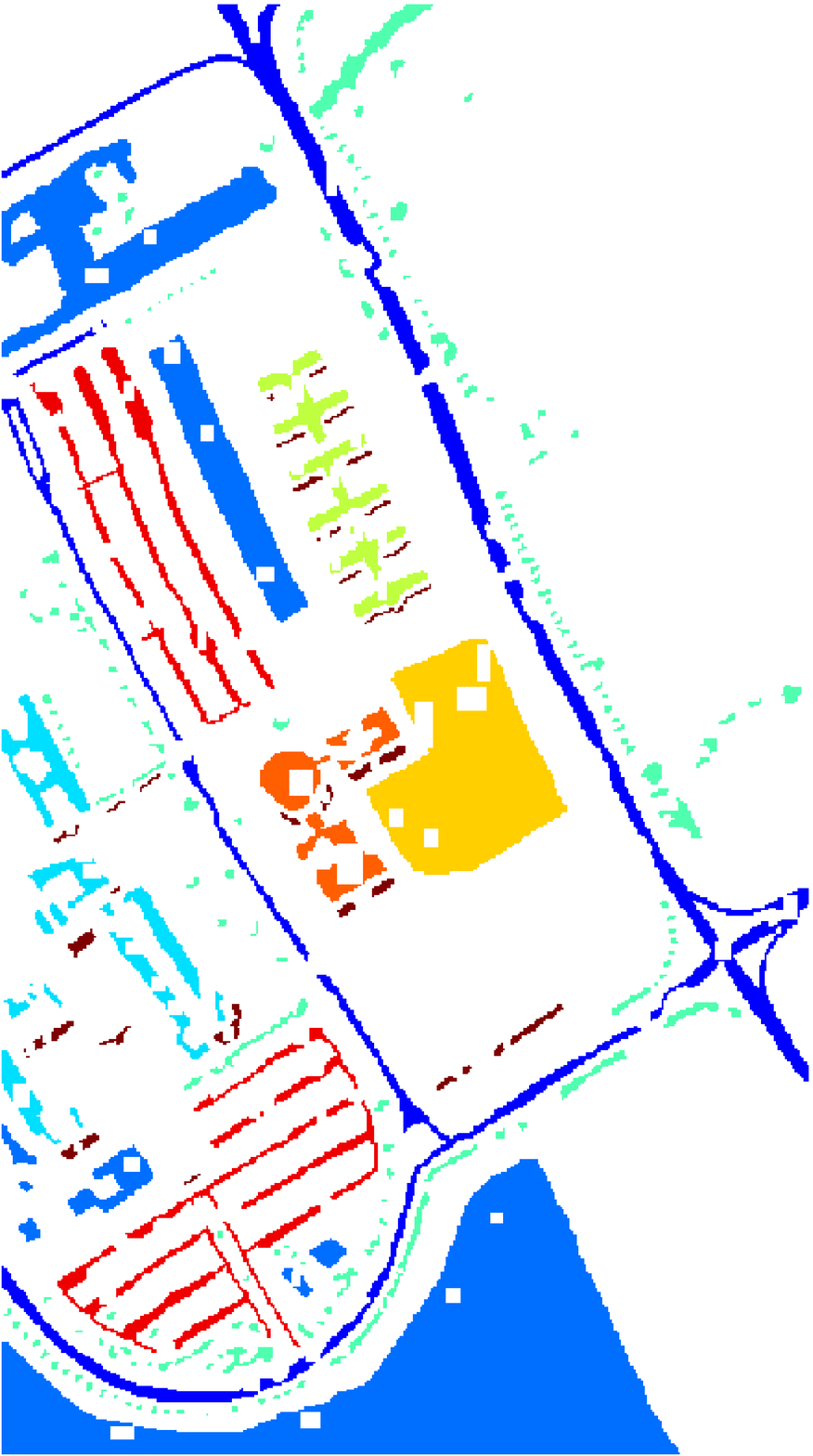}
\caption*{(b)}
\end{minipage}
%\hspace{0.3cm}
%\begin{minipage}[b]{0.3\linewidth}
%\centering
%\includegraphics[width=\textwidth]{upavia_gt.eps}
%\caption*{(b)}
%\end{minipage}

%\captionsetup{justification=left}
%\captionsetup{justification=raggedright, singlelinecheck=false}
\caption{(a) Training sets and (b) test sets of the University of Pavia image.}
\label{fig:upavia_samples}
\end{figure}

\begin{table}[ht]

      \caption{Number of training and test samples for the University of Pavia image}
      \centering    
        \begin{tabular}[ht]{ c| c || c | c  }
\hline
Class \#  &Name & Train & Test \\
\hline
 1    &Asphalt        &548 &6304 \\
 2  &Meadows          &540 &18146 \\
 3  &Gravel           &392 &1815   \\
 4  &Trees            &524 &2912 \\
 5  &Metal sheets     &265 &1113 \\
 6  &Bare soil        &532 &4572\\
 7  &Bitumen          &375 &981 \\
 8  &Bricks           &514 &3364\\
 9  &Shadows          &231 &795 \\
\hline
\multicolumn{2}{c||}{Total}  &3921 &40002\\
\hline

\end{tabular}
\label{table:university_pavia_sample}
\end{table}

\begin{figure*}[ht]

\centering
\begin{minipage}[b]{0.18\linewidth}
\centering
\includegraphics[width=\textwidth]{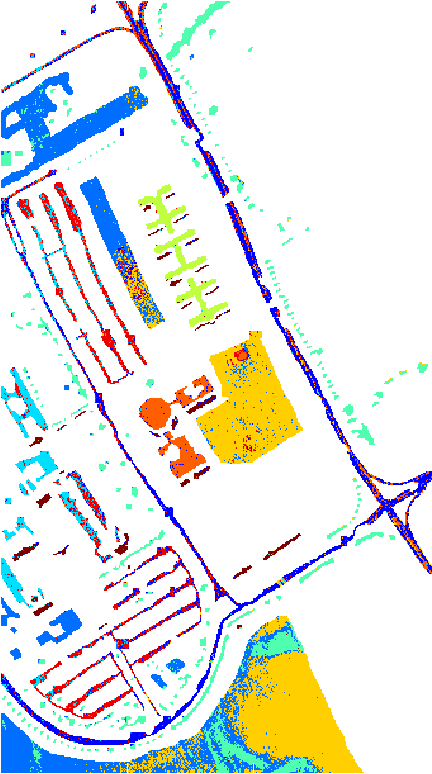}
\caption*{\tiny (a) SVM, OA = $69.84\%$}
\label{fig:figure2}
\end{minipage}
\hspace{0.1cm}
\begin{minipage}[b]{0.18\linewidth}
\centering
\includegraphics[width=\textwidth]{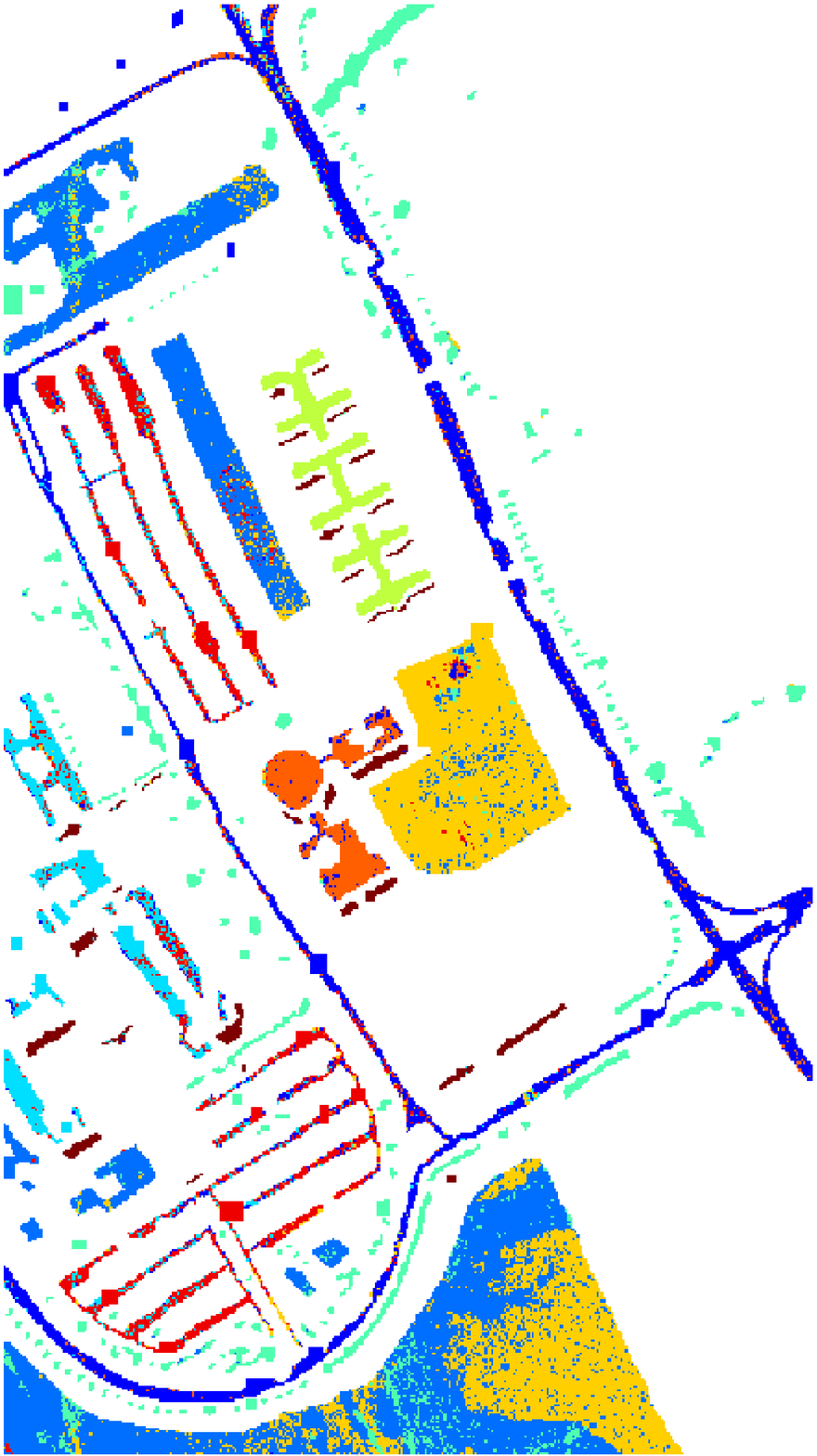}  %SRC
\caption*{\tiny (b) SRC, OA = $66.51\%$}
\label{fig:figure2}
\end{minipage}
\hspace{0.1cm}
\begin{minipage}[b]{0.18\linewidth}
\centering
\includegraphics[width=\textwidth]{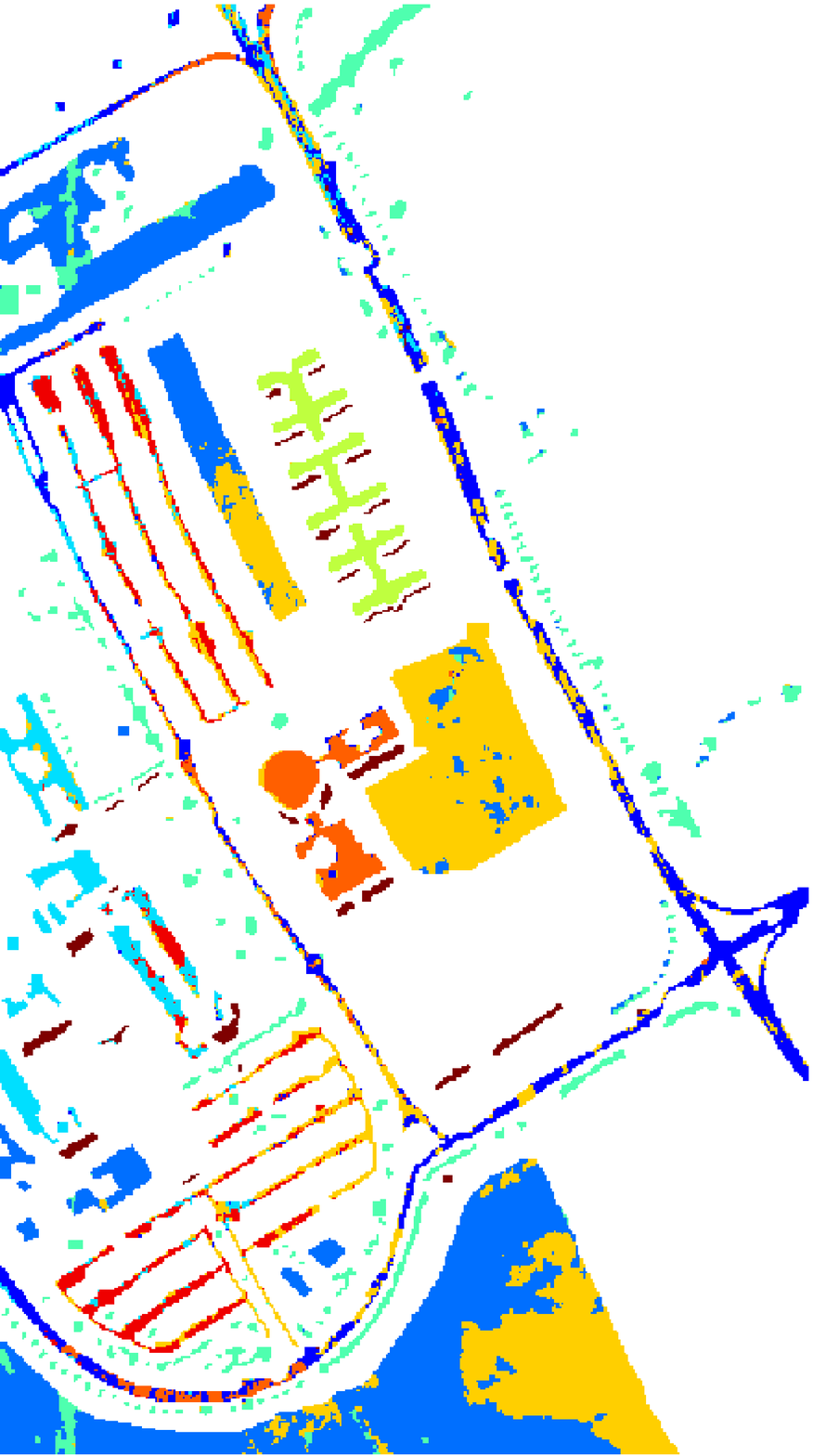}   %src-js
\caption*{\tiny (c) SRC-JS, OA = $74.05\%$}
\label{fig:figure2}
\end{minipage}
\hspace{0.1cm}
\begin{minipage}[b]{0.18\linewidth}
\centering
\includegraphics[width=\textwidth]{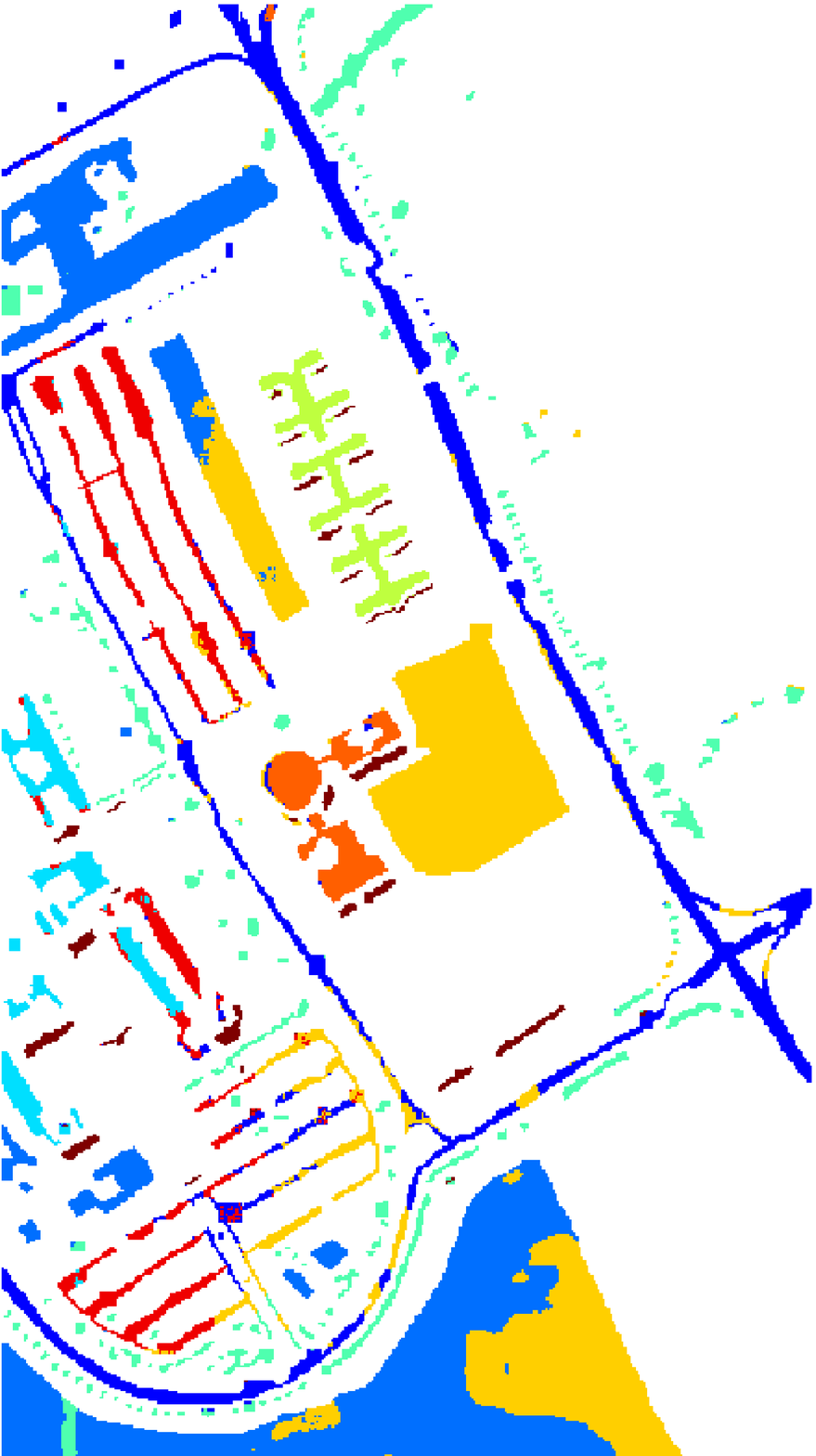}   %src-lp
\caption*{\tiny (d) SRC-LP, OA = $80.82\%$}
\label{fig:figure2}
\end{minipage}
\begin{minipage}[b]{0.18\linewidth}
\centering
\includegraphics[width=\textwidth]{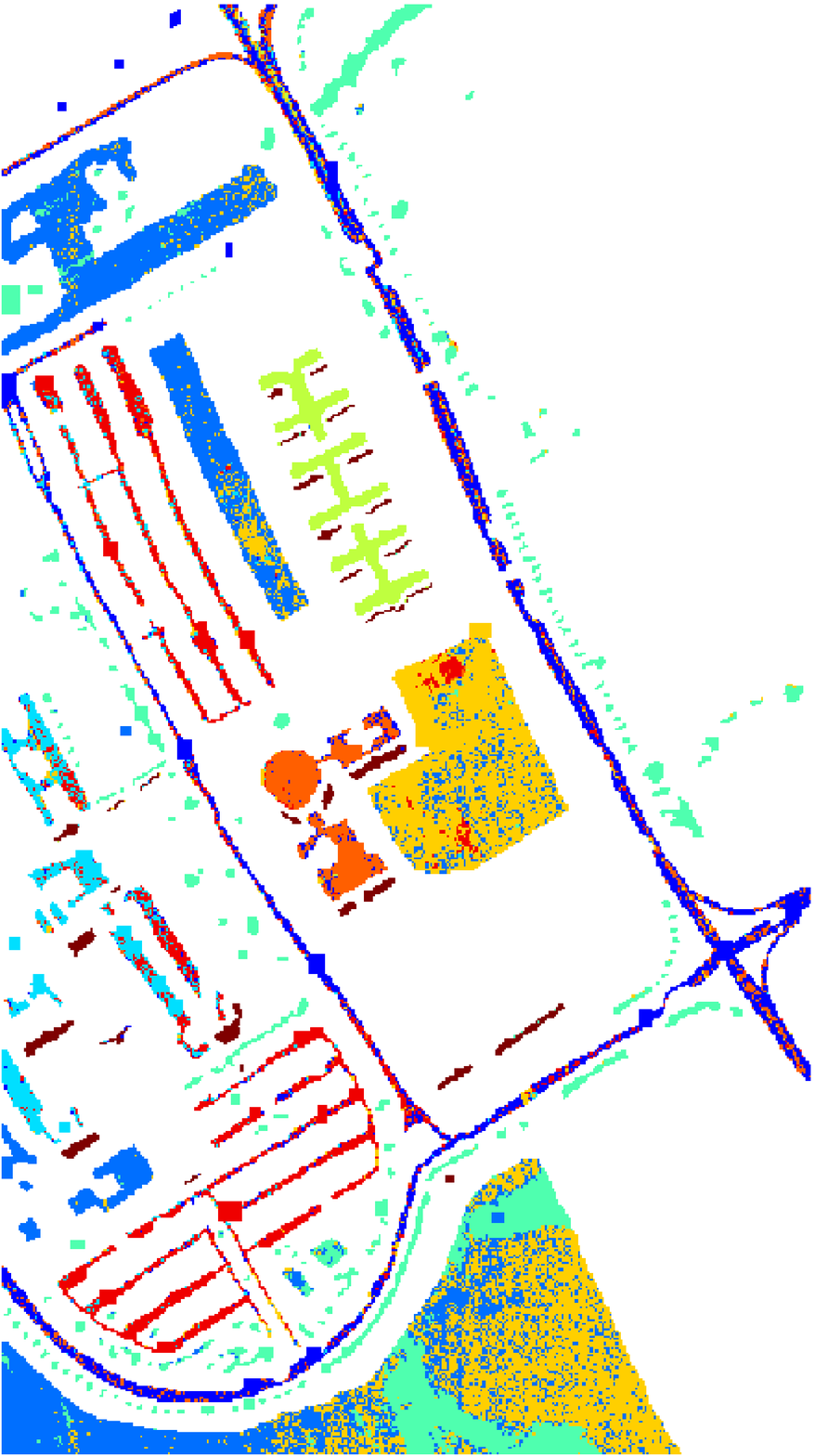}  %l1-unsup
\caption*{\tiny (e) ODL, OA = $64.57\%$}
\end{minipage}

\begin{minipage}[b]{0.18\linewidth}
\centering
\includegraphics[width=\textwidth]{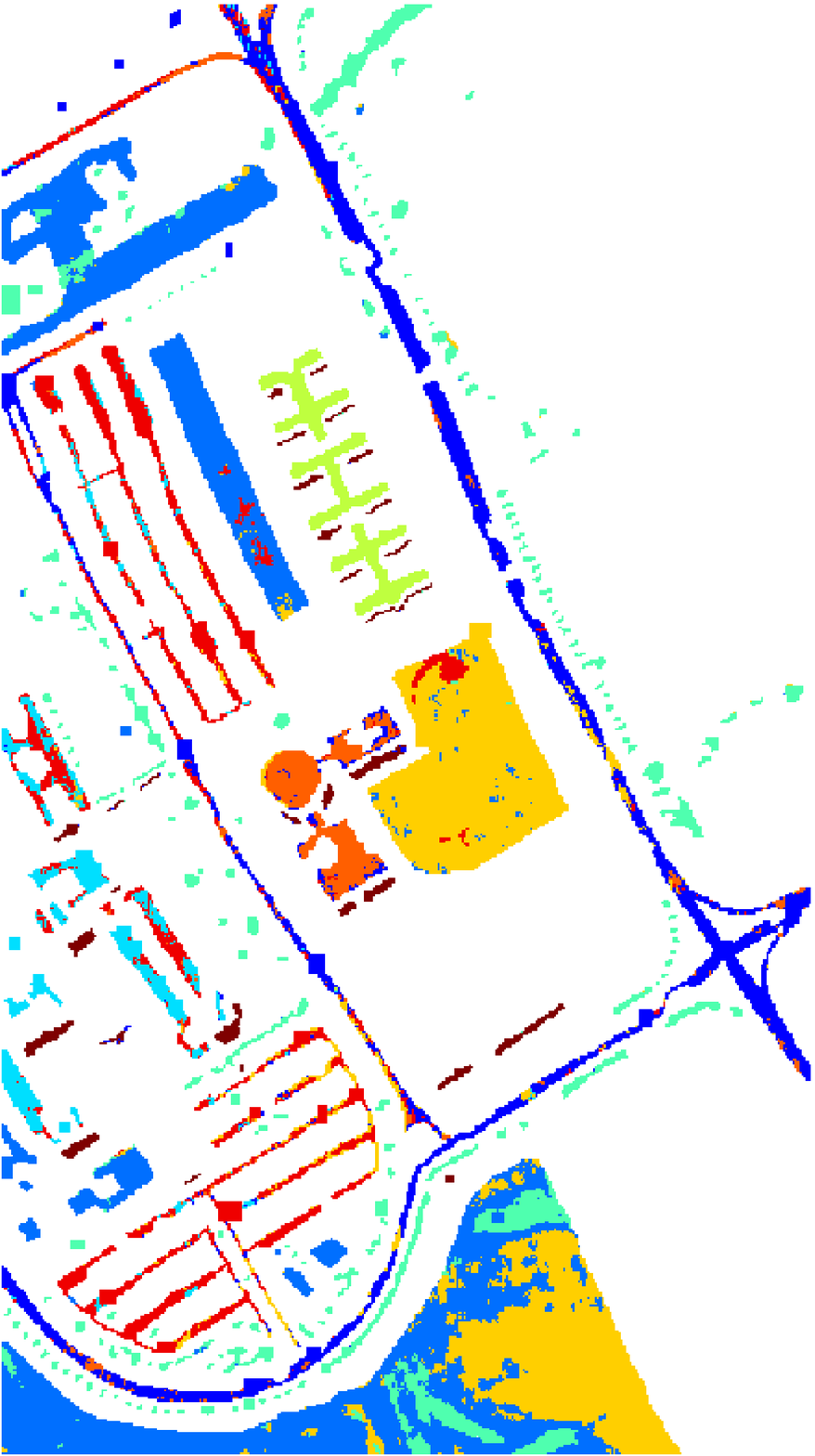}
\caption*{\tiny (f)  ODL-JS, OA = $75.83\%$}
\end{minipage}
\begin{minipage}[b]{0.18\linewidth}
\centering
\includegraphics[width=\textwidth]{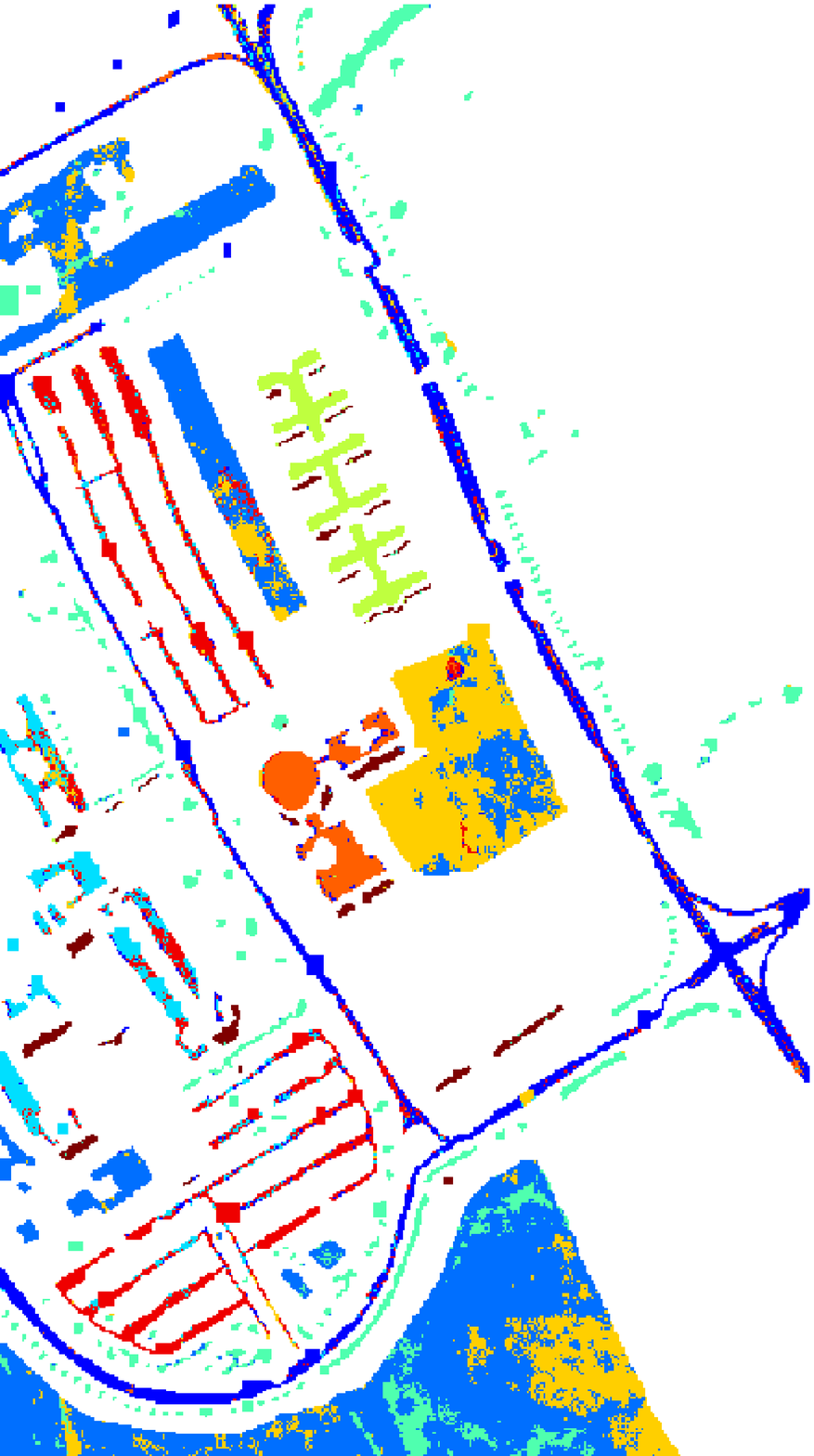}   %lp-unsup
\caption*{\tiny (g) ODL-LP, OA = $78.15\%$}
\label{fig:figure2}
\end{minipage}
\hspace{0.1cm}
\begin{minipage}[b]{0.18\linewidth}
\centering
\includegraphics[width=\textwidth]{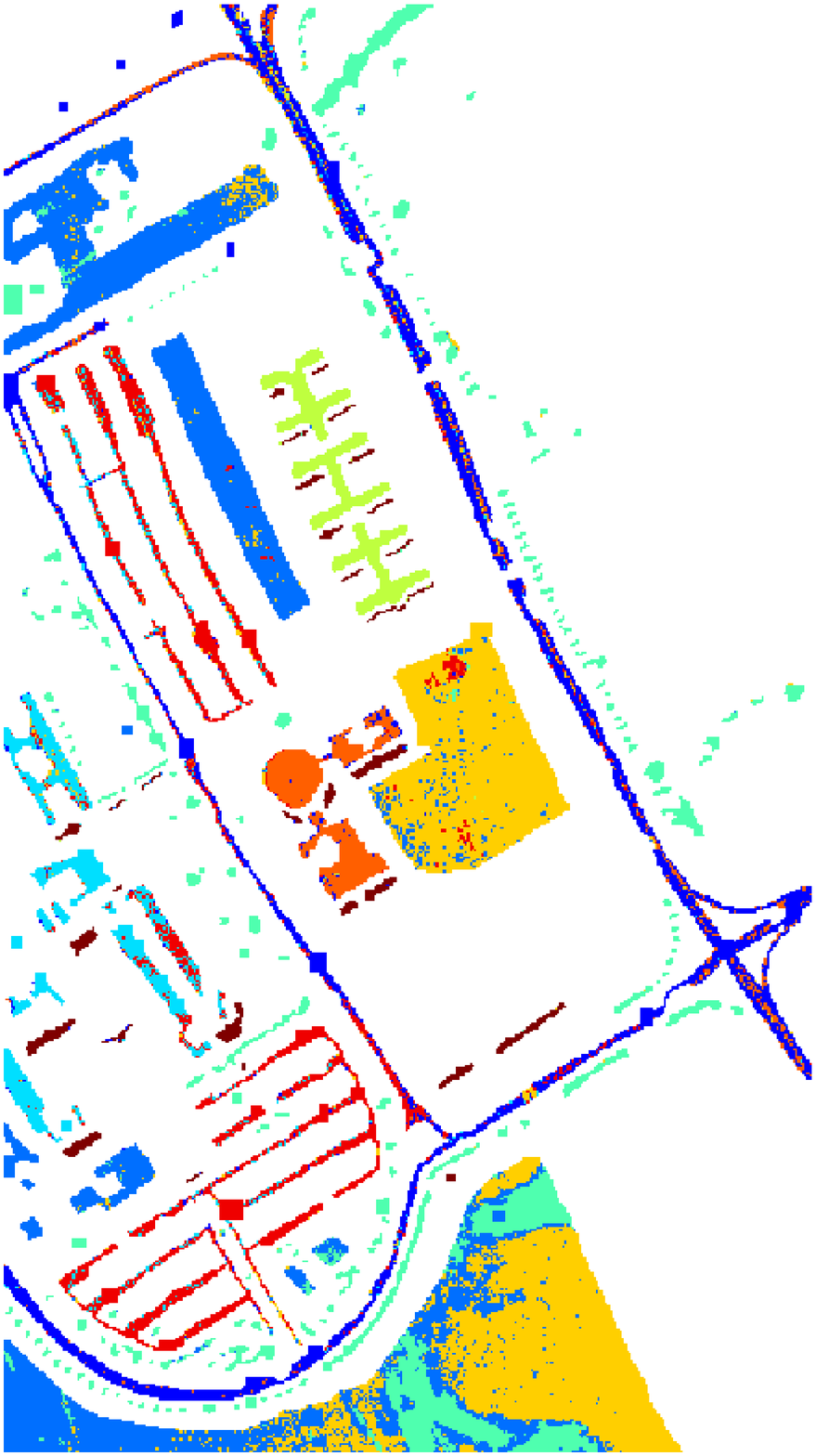}  %l1-sup
\caption*{\tiny (h) TDDL, OA = $69.30\%$}
\end{minipage}
\hspace{0.1cm}
\begin{minipage}[b]{0.18\linewidth}
\centering
\includegraphics[width=\textwidth]{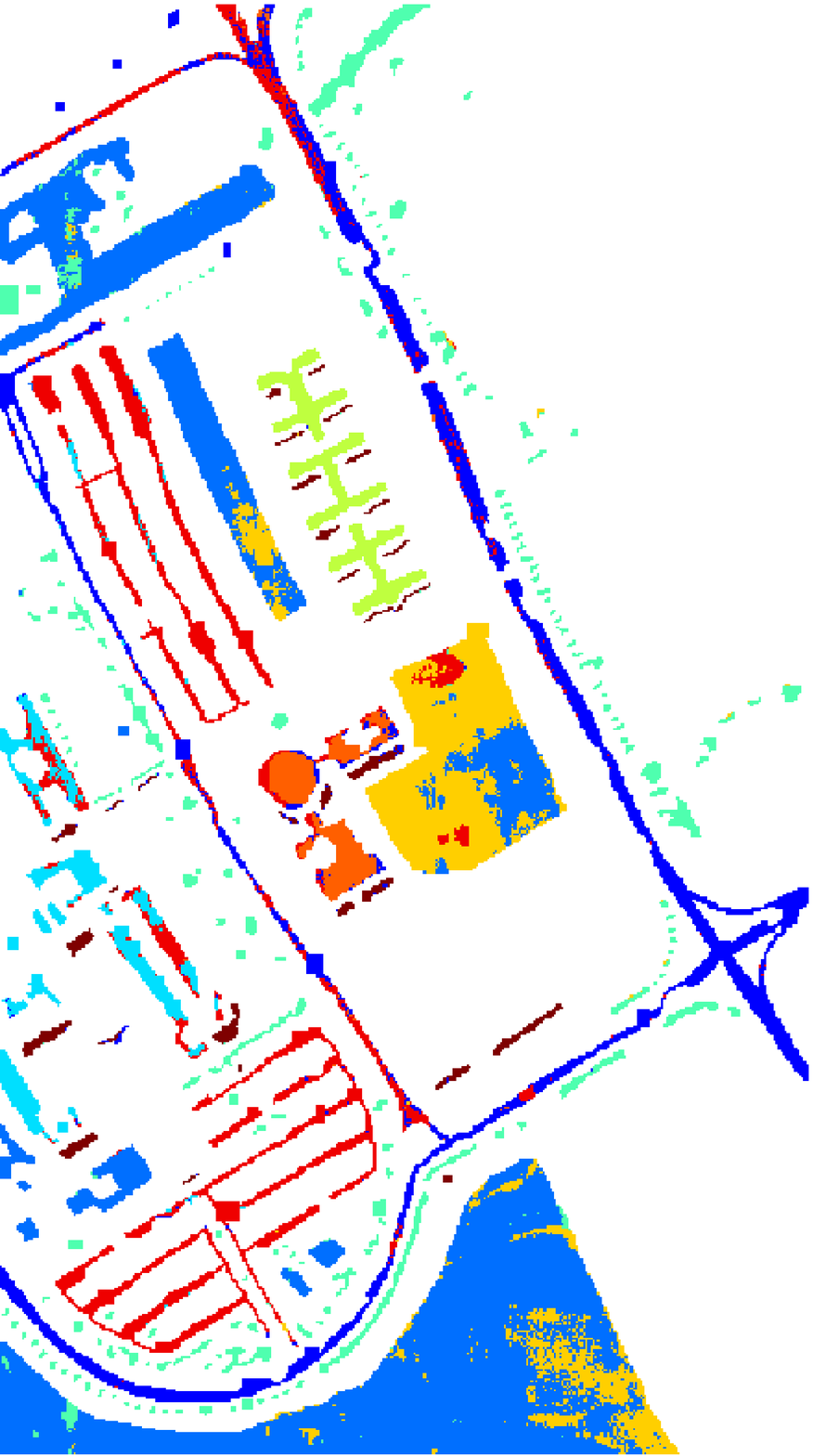}  %for js-sup
\caption*{\tiny (i)  TDDL-JS, OA = $84.48\%$}
\end{minipage}
\hspace{0.1cm}
\begin{minipage}[b]{0.18\linewidth}
\centering
\includegraphics[width=\textwidth]{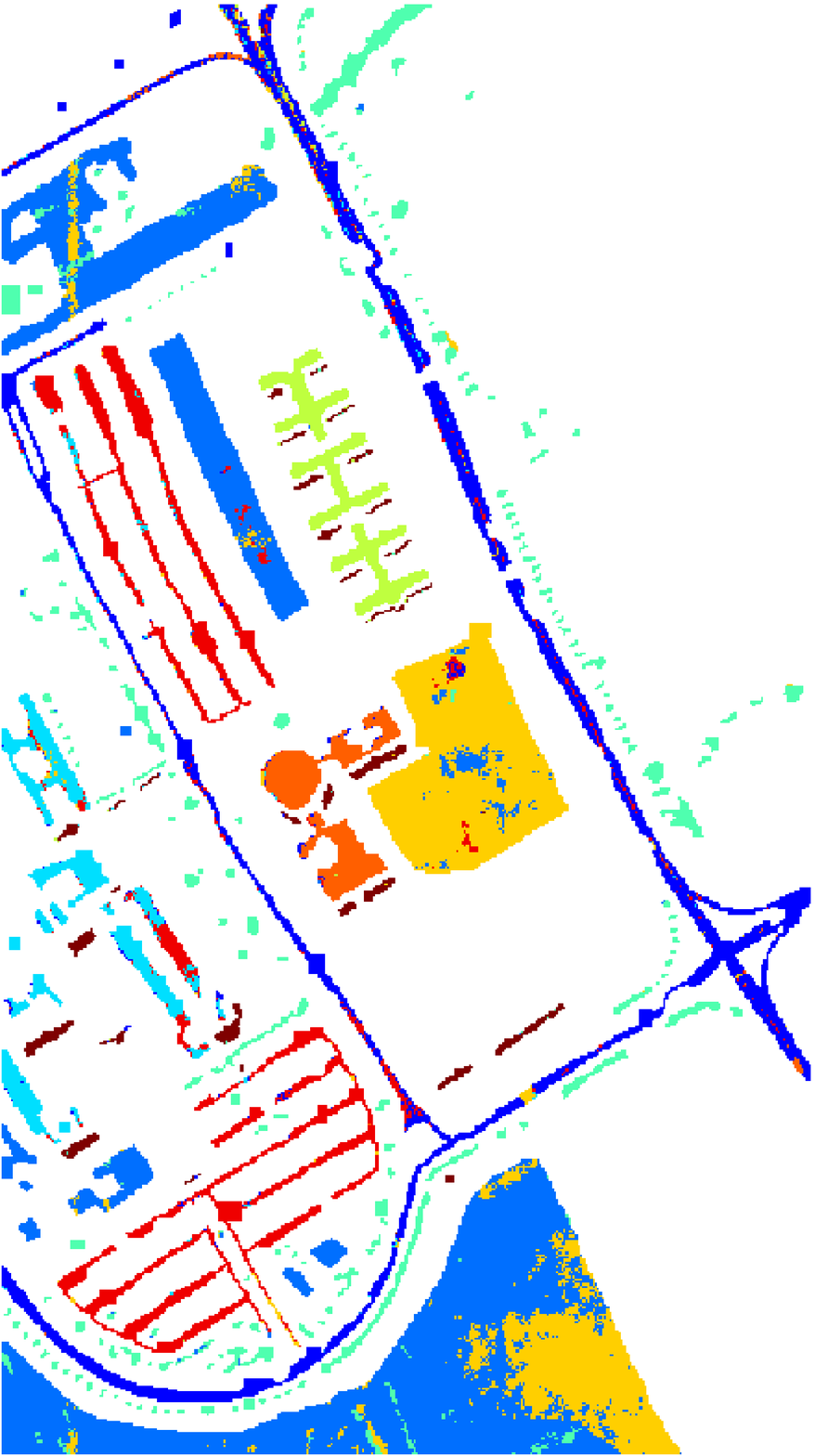}  %for LP sup
\caption*{\tiny (j) TDDL-LP, OA = $85.70\%$}
\label{fig:figure1}
\end{minipage}

%\captionsetup{justification=left}
\captionsetup{justification=raggedright, singlelinecheck=false}
\caption{Classification map of the University of Pavia image obtained by (a) SVM, (b) SRC, (c) SRC-JS, (d) SRC-LP, (e) ODL, (f) ODL-JS, (g) ODL-LP, (h) TDDL,  (i) TDDL-JS and (j) TDDL-LP.}
\label{fig:university_pavia}
\end{figure*}

\begin{table*}[ht]

      \centering
        \caption{Classification accuracy ($\%$) for the University of Pavia image}
\begin{tabular}[ht]{  c || c | c c c | c   c   c   c   c   c   }

%\hline
%\multicolumn{2}{|c|}{Optimization Techniques} &\multicolumn{6}{|c||}{ADMM/SpaRSA}\\
\hline
\multicolumn{2}{c|}{Dictionary Size} &\multicolumn{3}{c|}{$N=3921$} &\multicolumn{6}{c}{$N=45$}\\
\hline
Class    & SVM &SRC &SRC-JS &SRC-LP  & ODL & ODL-JS  &ODL-LP & TDDL & TDDL-JS  &TDDL-LP \\
\hline
 1  &84.55   &57.11 &77.04  &\textbf{95.08} &39.16 &86.64  &79.38  &74.60  &79.27 &87.77 \\
 2  &82.45 &58.22 &67.98  &66.70 &66.37 &56.48 &75.89 &51.27  &\textbf{86.85} &78.89 \\
 3  &77.08 &57.33 &44.32  &77.55 &65.40  &\textbf{80.72} &62.42 &77.19  &71.13 &78.79 \\
 4  &94.19 &95.94 &95.13 &95.19 &78.67 &\textbf{99.04} &96.91 &98.08  &98.87 &98.21 \\
 5  &99.01 &\textbf{100.00} &99.85  &\textbf{100.00}&99.91 &100.00 &99.82  &99.91  &99.91 &99.91 \\
 6  &23.55  &89.60 &88.31 &96.60 &64.94 &\textbf{96.89} &72.13 &90.07  &68.74 &91.64 \\
 7  &2.06  &83.27 &96.59 &\textbf{96.59} &91.64 &91.23 &84.10 &86.14  &68.09 &93.17 \\
 8  &33.89  &48.65 &65.20 &67.36&67.36 &90.81 &75.98 &78.00  &95.54 &\textbf{94.20} \\
 9  &53.05  &93.69 &99.59 &\textbf{99.59}&71.07 &98.37 &93.46 &95.72  &91.82 &95.09 \\

\hline
\textbf{OA}[$\%$]  &69.84  &66.51 &74.05 &80.82  &64.57 &75.83 &78.15 &69.30 &84.48 &\textbf{85.70} \\
\textbf{AA}[$\%$]  &61.09  &75.98 &80.06 &88.80 &71.66 &88.91 &82.23 &83.44 &84.47  &\textbf{90.85} \\
$\mathbf{\kappa}$  &0.569 &0.628 &0.681 &0.758 &0.549 &0.731 &0.747 &0.662 &0.817 &\textbf{0.835} \\
\hline

\end{tabular}
\label{table:upavia_result}
\end{table*}

\subsection{Classification on ROSIS Pavia Urban Data Set}
The last two images to be tested are the University of Pavia and the Center of Pavia, which are urban images acquired by the Reflective Optics System Imaging Spectrometer (ROSIS). It generates 115 spectral bands ranging from $0.43$ to $0.86\mu$m. 

The University of Pavia image contains $610\times340$ pixels.  12 noisiest bands out of all 115 bands are removed. There are nine ground-truth classes of interests as shown in Table \ref{table:university_pavia_sample}. For this image, the training samples were manually labelled by  an analyst. The total number of training and testing samples is $3,921$ ($10.64\%$  of all the interested pixels) and $40,002$ respectively. The training and testing map are visually displayed in Fig. \ref{fig:upavia_samples}.

 For the University of Pavia, we set the total iterations of unsupervised and supervised dictionary learning methods to be $30$ and $200$ respectively.  The window size is set to $5\times 5$ for all joint or Laplacian sparse regularized methods to obtain the highest OA. The ODL-LP is able to reach a performance of $78.15\%$ for OA, which is more than $14\%$ higher than that of ODL. The ODL-JS also significantly improves the OA, which is more than $11\%$ higher than that of ODL. TDDL-LP has the highest OA = $85.70\%$, which indicates that it  outperforms other methods when classify large regions of the image. It also has the highest $\kappa = 0.935$. The best classification accuracy for class $1$ (Asphalt), which consists of narrow strips, is obtained by using TDDL-LP ($87.77\%$). Class 2 (Meadows) is composed of large smooth regions, as expected, TDDL-JS gives the highest accuracy ($86.85\%$) for this class. TDDL has large amount of misclassification pixels for class 2. The highest AA ($90.85\%$) is given by TDDL-LP, which confirms that the TDDL-LP is superior to other methods when classify the pixels in non-homogeneous regions. 

\begin{figure}[!t]
\centering

\begin{minipage}[b]{0.35\linewidth}
\centering
\includegraphics[width=\textwidth]{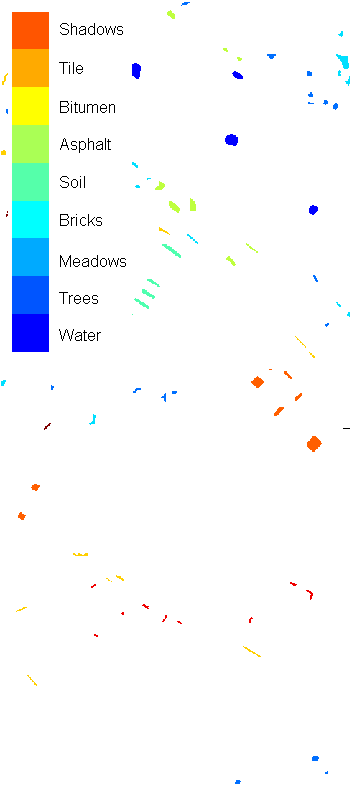}
\caption*{(a)}
\end{minipage}
\hspace{0.3cm}
\begin{minipage}[b]{0.35\linewidth}
\centering
\includegraphics[width=\textwidth]{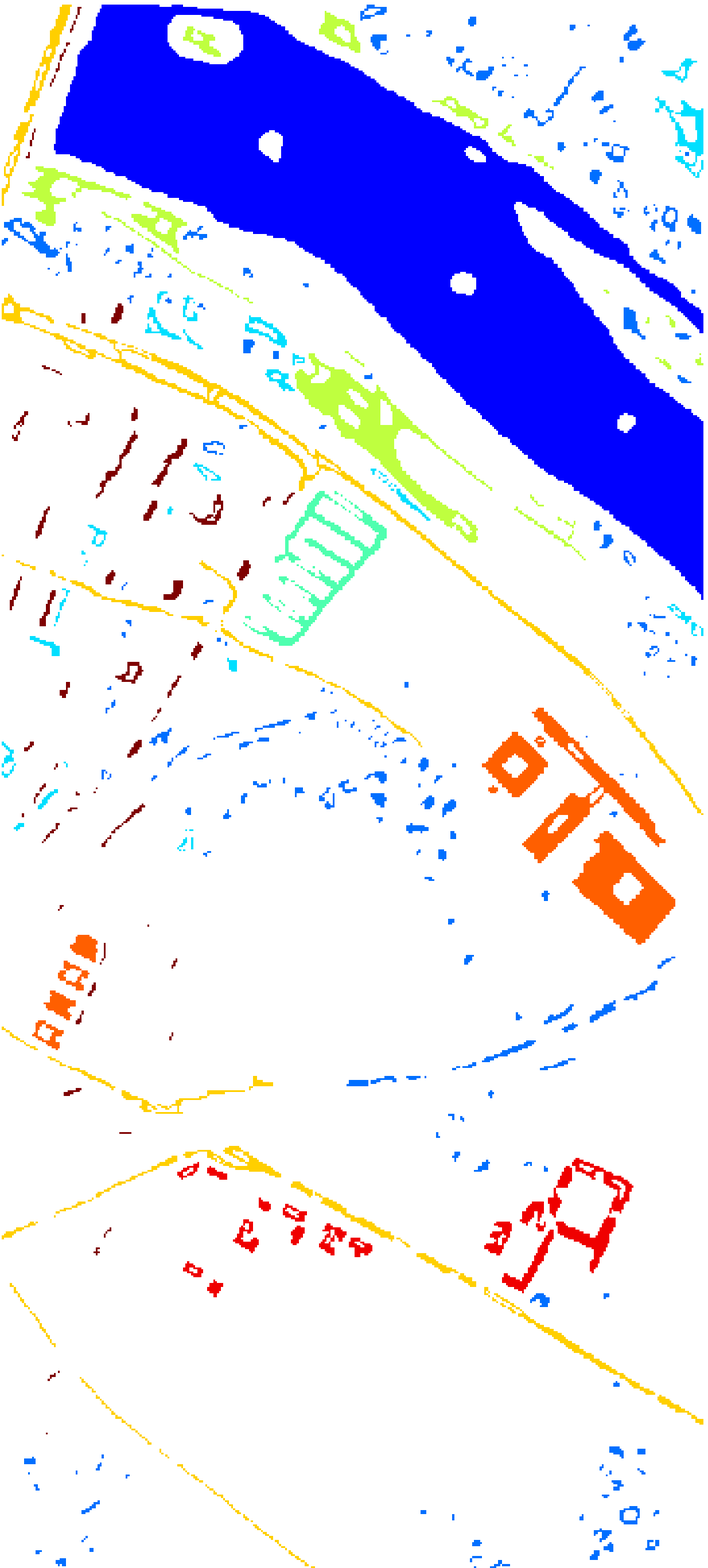}
\caption*{(b)}
\end{minipage}

%\captionsetup{justification=left}
%\captionsetup{justification=raggedright, singlelinecheck=false}
\caption{(a) Training sets and (b) test sets of the Center of Pavia image.}
\label{fig:cpavia_samples}
\end{figure}

The third image where we evaluate various approaches is the Center of Pavia, which consists of $1094\times 492$ pixels. Each pixel has $102$ bands after removing $13$ noisy bands. This image consists of nine ground-truth classes of interest as shown in Table \ref{table:center_pavia_sample} and Fig. \ref{fig:cpavia_samples}. $5,536$ manually labelled pixels are designated as the training samples and the remaining $97,940$ interested pixels are used for testing.

\begin{table}[!]

      \caption{Number of training and test samples for the Center of Pavia image}
      \centering    
        \begin{tabular}[ht]{ c| c || c | c  }
\hline
Class \#  &Name & Train & Test \\
\hline
 1    &Water        &745 &64533 \\
 2  &Trees          &785 &5722 \\
 3  &Meadows           &797 &2094   \\
 4  &Bricks            &485 &1667 \\
 5  &Soil     &820 &5729 \\
 6  &Asphalt        &678 &6847 \\
 7  &Bitumen          &808 &6479 \\
 8  &Tile           &223 &2899 \\
 9  &Shadows      &195 &1970 \\
\hline
\multicolumn{2}{c||}{Total}  &5536 &97940 \\
\hline

\end{tabular}
\label{table:center_pavia_sample}
\end{table}

\begin{figure*}[ht]

\hspace{0.1cm}
\begin{minipage}[b]{0.18\linewidth}
\centering
\includegraphics[width=\textwidth]{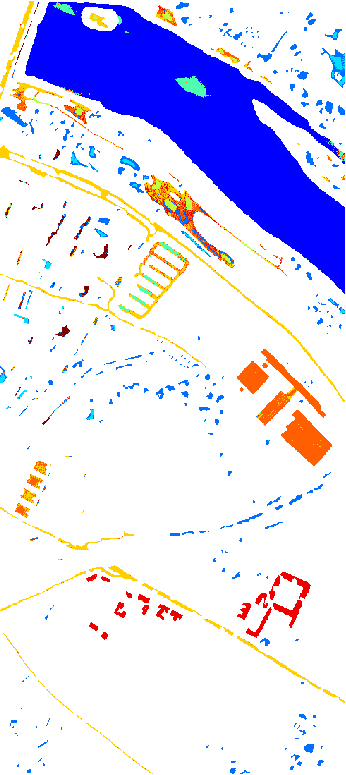}
\caption*{\tiny (a) SVM, OA = $95.68\%$}
\label{fig:figure2}
\end{minipage}
\hspace{0.1cm}
\begin{minipage}[b]{0.18\linewidth}
\centering
\includegraphics[width=\textwidth]{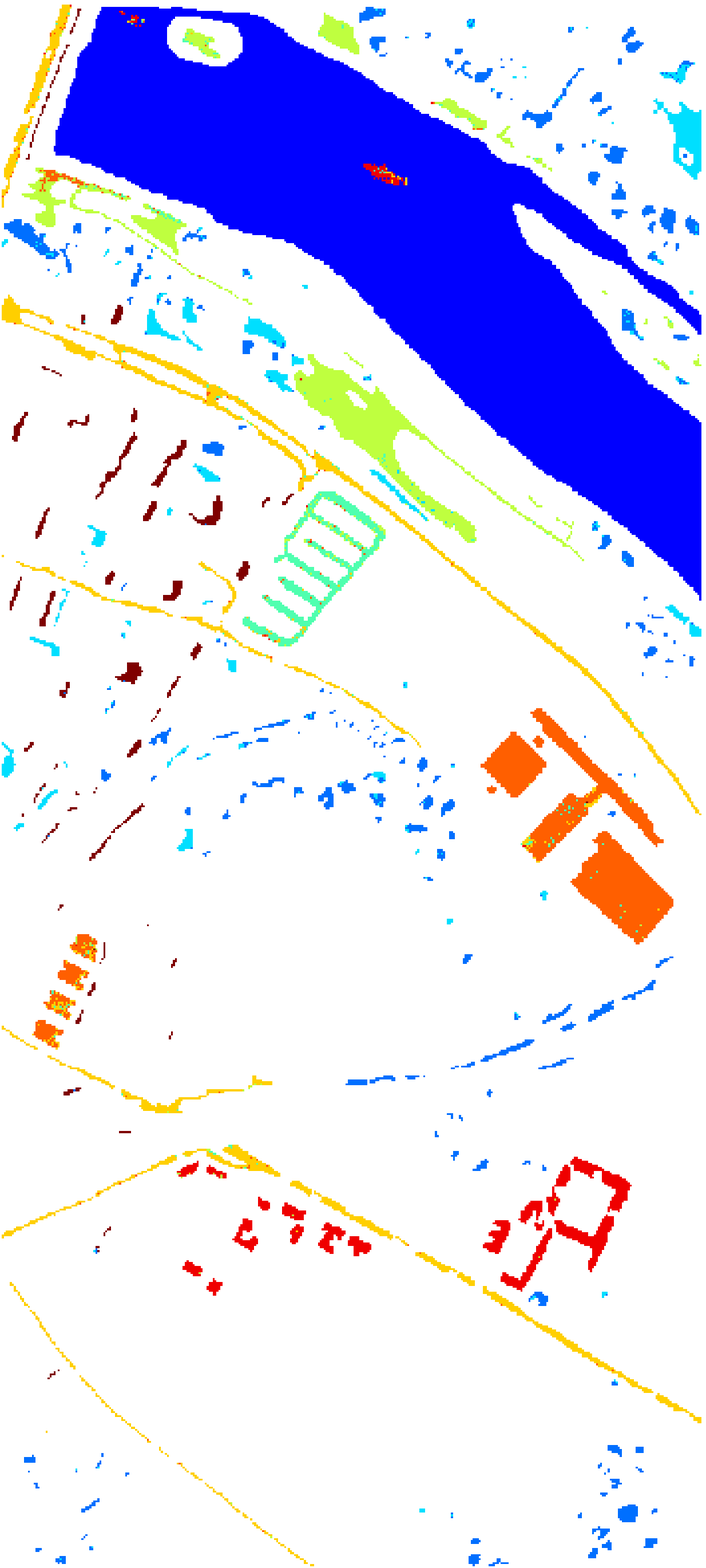} %SRC
\caption*{\tiny (b) SRC, OA = $97.57\%$}
\label{fig:figure2}
\end{minipage}
\hspace{0.1cm}
\begin{minipage}[b]{0.18\linewidth}
\centering
\includegraphics[width=\textwidth]{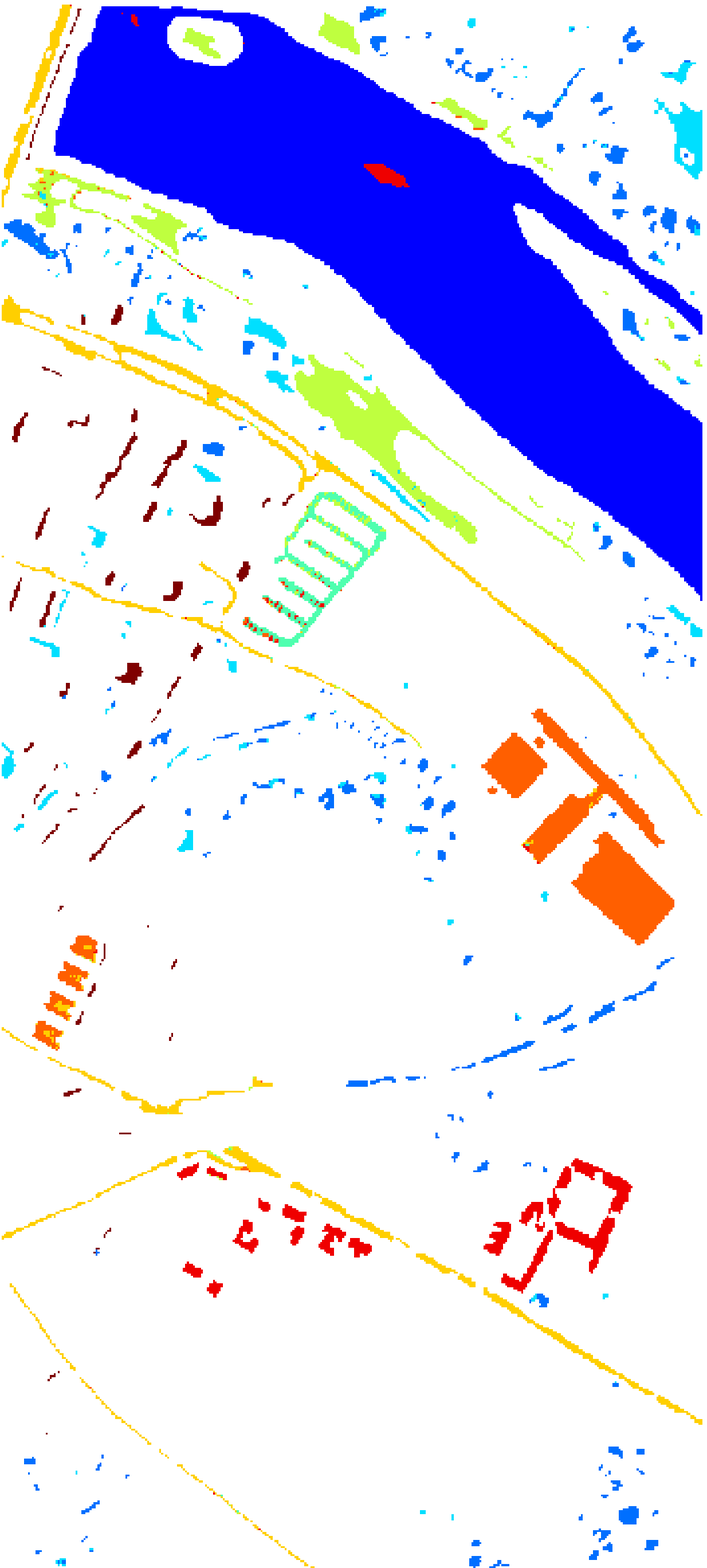}   %SRC-js
\caption*{\tiny (c)  SRC-JS, OA = $98.01\%$}
\label{fig:figure2}
\end{minipage}
\hspace{0.1cm}
\begin{minipage}[b]{0.18\linewidth}
\centering
\includegraphics[width=\textwidth]{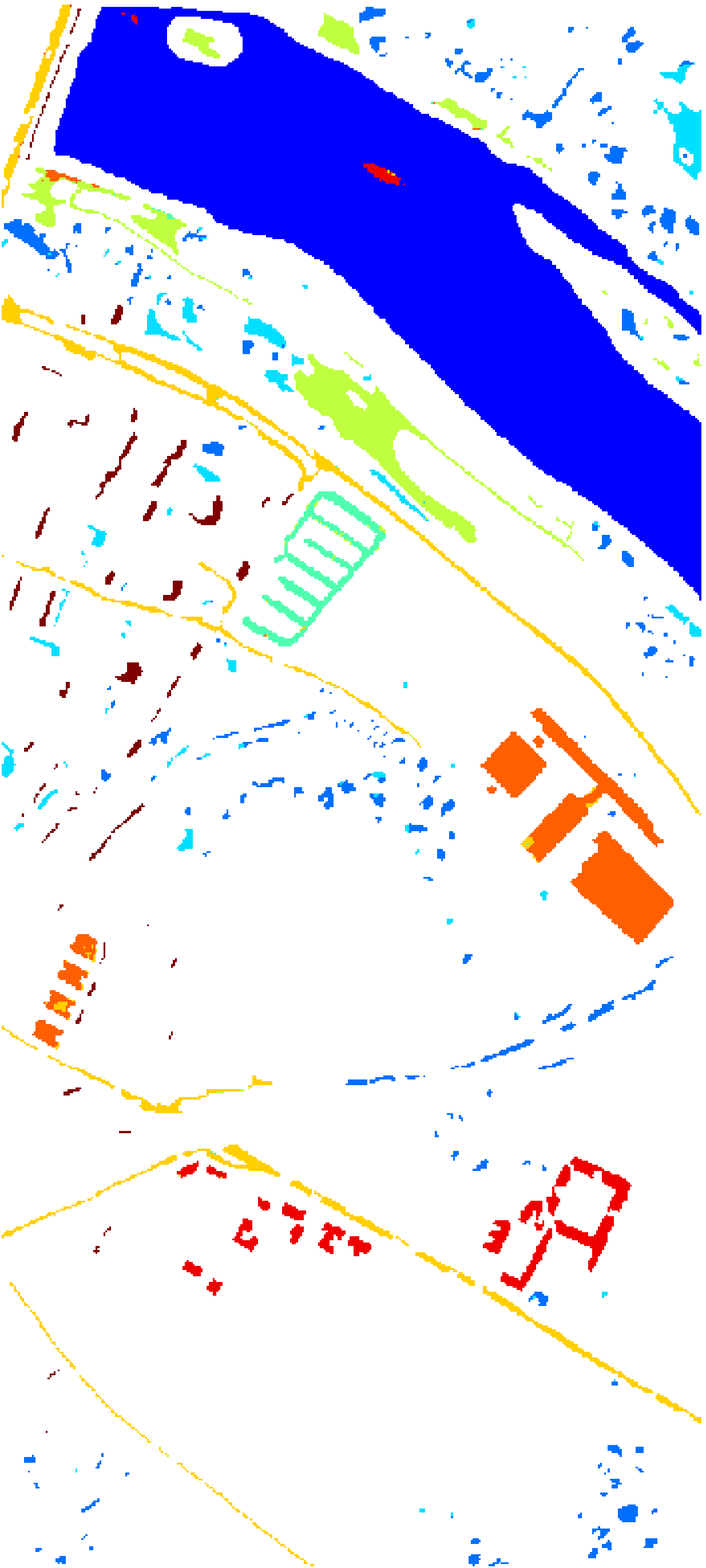} %SRC-lp
\caption*{\tiny (d) SRC-LP, OA = $98.36\%$}
\label{fig:figure2}
\end{minipage}
\begin{minipage}[b]{0.18\linewidth}
\centering
\includegraphics[width=\textwidth]{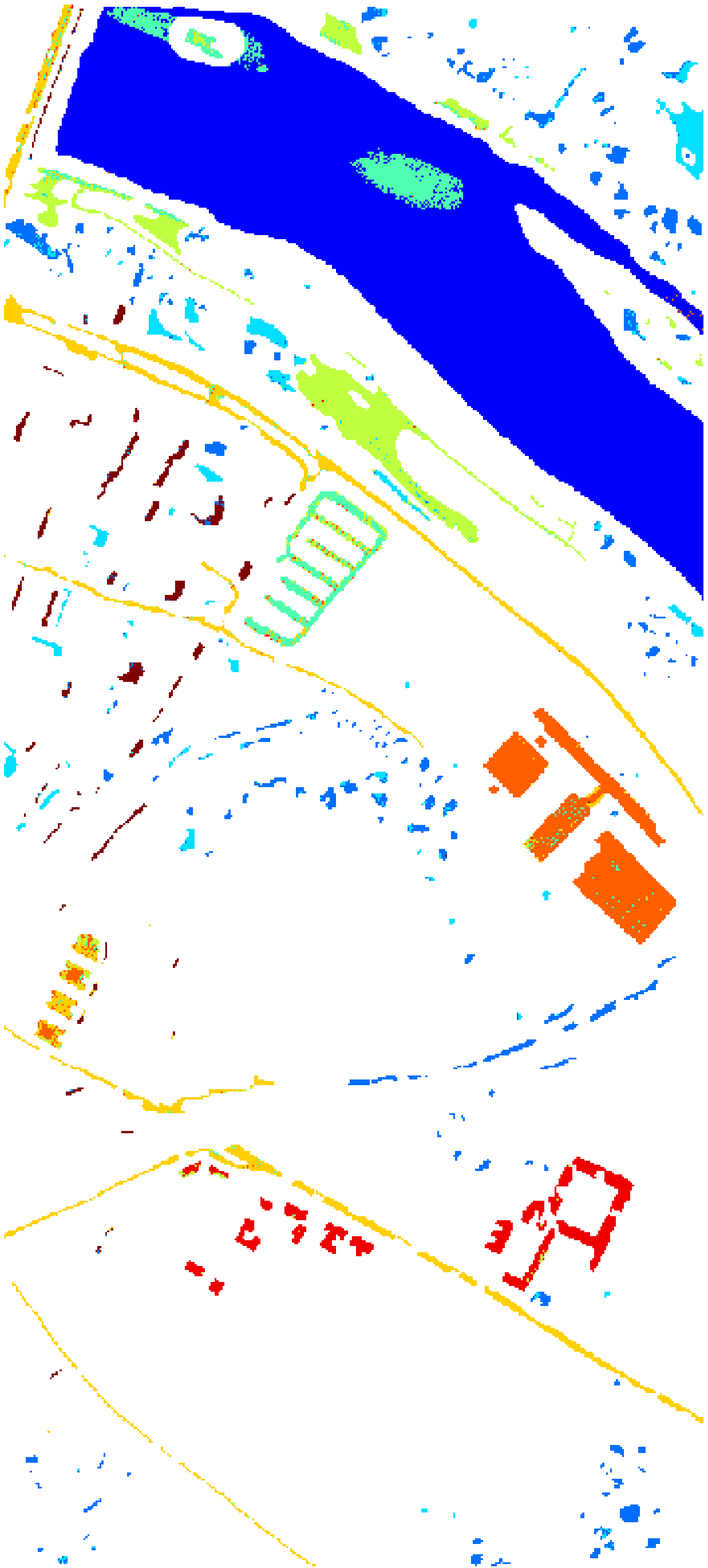}  %unsup-l1
\caption*{\tiny (e) ODL, OA = $93.67\%$}
\end{minipage}

\begin{minipage}[b]{0.18\linewidth}
\centering
\includegraphics[width=\textwidth]{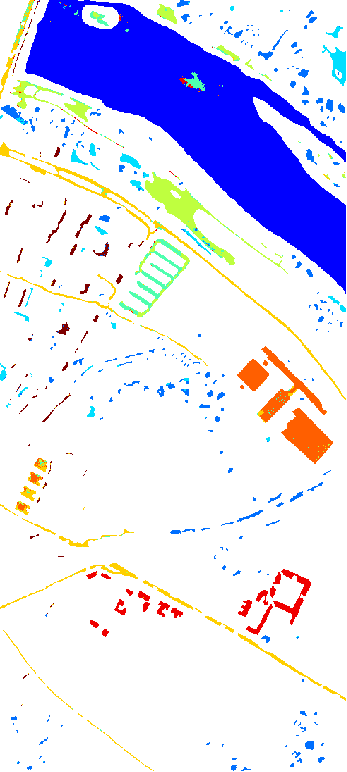}
\caption*{\tiny (f) ODL-JS, OA = $96.13\%$}
\end{minipage}
\begin{minipage}[b]{0.18\linewidth}
\centering
\includegraphics[width=\textwidth]{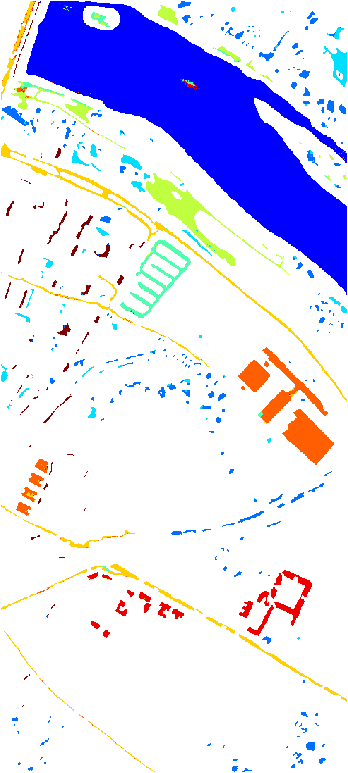}   %unsup-lp
\caption*{\tiny (g) ODL-LP, OA = $97.86\%$}
\label{fig:figure2}
\end{minipage}
\hspace{0.1cm}
\begin{minipage}[b]{0.18\linewidth}
\centering
\includegraphics[width=\textwidth]{cpavia_l1_unsup.eps}  %sup-l1
\caption*{\tiny (h) TDDL, OA = $96.30\%$}
\end{minipage}
\hspace{0.1cm}
\begin{minipage}[b]{0.18\linewidth}
\centering
\includegraphics[width=\textwidth]{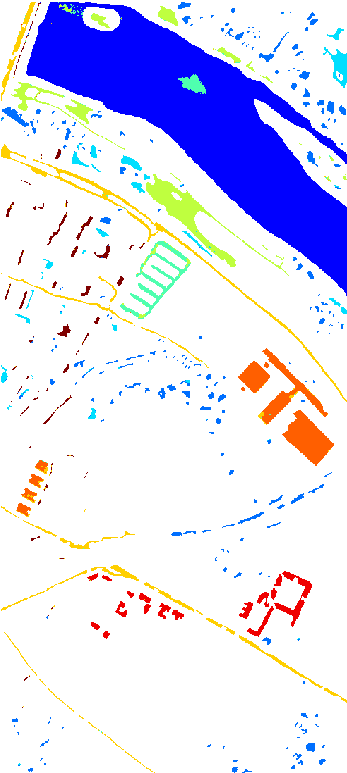}  %for LP-SRC
\caption*{\tiny (i) TDDL-JS, OA = $98.01\%$}
\end{minipage}
\hspace{0.1cm}
\begin{minipage}[b]{0.18\linewidth}
\centering
\includegraphics[width=\textwidth]{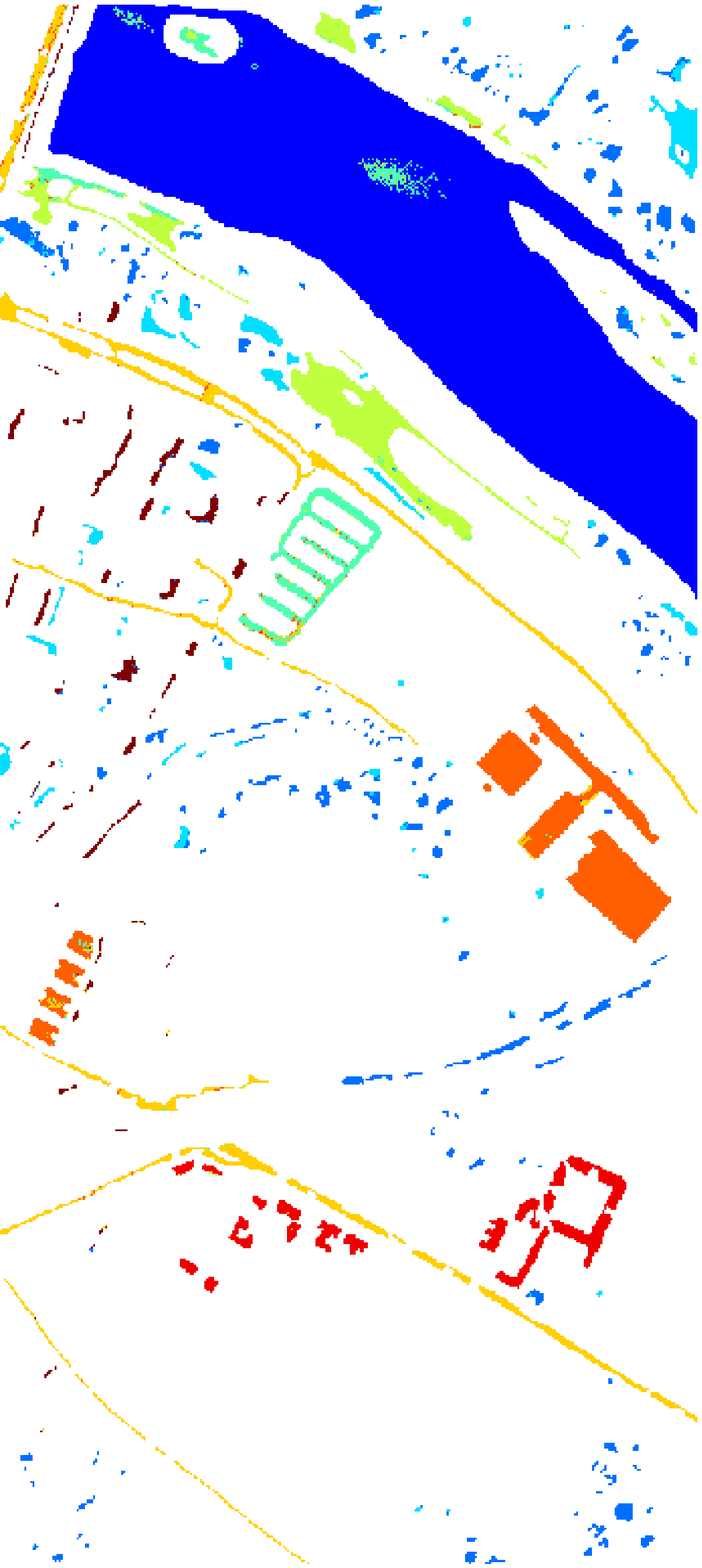}
\caption*{\tiny (j) TDDL-LP, OA = $98.67\%$}
\label{fig:figure1}
\end{minipage}

%\captionsetup{justification=left}
\captionsetup{justification=raggedright, singlelinecheck=false}
\caption{Classification map of the Center of Pavia image obtained by (a) SVM, (b) SRC, (c) SRC-JS, (d) SRC-LP, (e) ODL, (f) ODL-JS, (g) ODL-LP, (h) TDDL,  (i) TDDL-JS and (j) TDDL-LP.}
\label{fig:center_pavia}
\end{figure*}

\begin{table*}

      \centering
        \caption{Classification accuracy ($\%$) for the Center of Pavia image}
\begin{tabular}[h]{  c || c | c c c | c   c   c   c   c   c   }

%\hline
%\multicolumn{2}{|c|}{Optimization Techniques} &\multicolumn{6}{|c||}{ADMM/SpaRSA}\\
\hline
\multicolumn{2}{c|}{Dictionary Size} &\multicolumn{3}{c|}{$N=5536$} &\multicolumn{6}{c}{$N=45$}\\
\hline
Class    & SVM &SRC &SRC-JS &SRC-LP  & ODL & ODL-JS  &ODL-LP & TDDL & TDDL-JS  &TDDL-LP \\
\hline
 1  &96.97 &99.58  &99.52 &99.28  &96.26 &99.13  &\textbf{99.69} &98.54  &98.76  &99.13 \\
 2  &91.09 &90.07  &96.89 &92.11  &84.25 &94.63  &90.63 &89.55  &\textbf{97.59}  &93.01 \\
 3  &96.08 &95.42  &99.47 &98.62  &93.36 &96.23  &97.61 &95.18  &96.85  &\textbf{98.71} \\
 4  &86.32 &79.96  &78.28 &94.72  &61.61 &64.73  &97.30 &85.78  &85.18  &\textbf{97.41} \\
 5  &88.57 &93.70  &97.05 &97.14  &89.40 &84.62  &90.00 &88.08  &98.25  &\textbf{99.59} \\
 6  &95.27 &95.62  &98.19 &97.18  &94.35 &95.03  &94.49 &94.39  &\textbf{99.36}  &99.18 \\
 7  &94.03 &93.86  &97.01 &96.84  &86.31 &86.90  &97.33 &91.65  &94.46  &\textbf{98.66} \\
 8  &99.83 &99.17  &99.66 &99.66  &96.76 &99.79  &99.00 &98.17  &99.38  &\textbf{99.73} \\
 9  &85.74 &98.58  &99.19 &\textbf{99.95}  &93.25 &90.56  &94.42 &95.53  &91.27  &95.61 \\

\hline
\textbf{OA}[$\%$]  &95.68 &97.57  &98.01 &98.36 &93.67  &96.13   &97.86 &96.30  &98.01 &\textbf{98.67} \\
\textbf{AA}[$\%$]  &93.77 &94.00  &95.03 &97.28 &88.39  &90.18  &95.61 &92.99  &95.68 &\textbf{97.89} \\
$\mathbf{\kappa}$  &0.923 &0.961  &0.965 &0.971 &0.899  &0.938 &0.965 &0.940  &0.968 &\textbf{0.979} \\
\hline

\end{tabular}
\label{table:cpavia_result}
\end{table*}

Since this image has more labeled samples  than the other two images, we set the total iterations of unsupervised and supervised dictionary learning methods to be $75$ and $1000$ respectively. The window size is set to $5\times 5$ for the joint sparse and Laplacian sparse regularized methods. Although the OA of most methods are close, the OA of ODL-JS and ODL-LP are still around $3\%$ higher than that of ODL. The TDDL-LP reach the highest OA = $98.67\%$ over all the other methods. The OA of TDDL-JS ($98.01\%$) is slightly lower  than that of the TDDL-LP. We notice that SRC-JS (OA = $98.01\%$) and SRC-LP (OA=$98.36\%$) also render competitive performance when compared to TDDL-JS and TDDL-LP due to the fact that the raw spectral features of this image is already highly discriminative. TDDL-LP outperforms other methods on almost all classes and works especially well for Class $4$ (Bricks), achieving highest accuracy of $97.41\%$. Except for SRC-LP where the accuracy is $94.72\%$, none of others reaches accuracy over $90\%$ for Class $4$. Additionally, the AA of TDDL-LP ($97.21\%$) is almost $2\%$ better than that of TDDL-JS ($95.68\%$). These results support our assertion that the Laplacian sparsity prior provides stronger discriminability on nonhomogeneous regions.  Performance comparison between the SRC-based  and TDDL-based methods have shown that the dictionary size can be drastically decreased by applying supervised dictionary learning while achieving even better performance.

\section{CONCLUSION}
\label{sec:conclusion}

In this paper, we proposed novel a task driven dictionary learning method with joint or Laplacian sparsity prior for HSI classification. The corresponding optimization algorithms are developed using fixed point differentiation, and are further simplified for ease of implementation. We also derived the optimization algorithm for solving the  Laplacian sparse recovery problem using SpaRSA, which improves the computational efficiency due to the availability of a more accurate descent direction. The performance and the behavior of the proposed methods, i.e. TDDL-JS and TDDL-LP, have been extensively studied on the popular hyperspectral images. The results confirm that both TDDL-JS and TDDL-LP give plausible results on smooth homogeneous regions, while TDDL-LP one works better for classifying small narrow regions.  Compared to TDDL-JS, TDDL-LP is able to obtain a more stable performance by describing the similarities of neighboring pixels' sparse codes more delicately. The results also confirm that a significantly better performance can still be achieved when joint or Laplacian prior is imposed by using a very small dictionary. The overall accuracy of our algorithm can be improved by applying kernelization to the proposed approach. This can be achieved by kernelizing the sparse representation \cite{GCampsValls} and using a composite kernel classifier \cite{kernel_yichen}.

\section{Appendix A}
\label{append}
%\numberwithin{equation}{A1}
We can  infer from Eq. (\ref{eq:diff_js_5}) that $vec\left({\partial \bA}/{\partial D_{mn} } \right) = \mathbf{0}$, $\forall n\in \Lambda^c$, which indicates ${\partial \calL} / {\partial D_{mn}} = \mathbf{0}$, $\forall n\in \Lambda^c$. Therefore, we only need to take the gradient ${\partial \calL}/{\partial D_{mn}}$, $\forall n\in \Lambda$ into account. 

From the Eq. (\ref{eq:chain_js}) and Eq. (\ref{eq:diff_js_5}), we achieve  the gradient for every element of $\tbD$,
\begin{align}
\label{eq:append_a_1}
 \frac{\partial \calL}{\partial \tD_{mn}} = vec\left( \frac{\partial \calL}{\partial \tbA} \right)^\top  \cdot  vec\left( \frac{\partial \tbA}{\partial \tD_{mn}} \right), 
 \end{align}
where $m= 1, \dots, M$ and $n=1,\dots, N_\Lambda$. Let $\bg = vec\left(\frac{\partial f}{\partial \tbA^\top} \right)= vec \left( \left( \bW \hat{\bA} - \hat{\bY} \right)^\top \tbW \right)$ and $\tbW = \bW^{\Lambda}$ is the $\Lambda$ columns of $\bW$. Expand Eq. (\ref{eq:diff_js_5}) and combine it with Eq. (\ref{eq:append_a_1}), we have
\begin{align}
\frac{\partial \calL}{\partial \tD_{mn}}  = U_{mn} - V_{mn} \mbox{ and } \frac{\partial \calL}{\partial \tbD}  = \bU - \bV,
\end{align}
where $\bU, \bV \in\bbR^{m\times N_\Lambda}$ and every element $U_{mn}, V_{mn}$ are defined as
\begin{align*}
U_{mn} & = \bg^\top\left( \tbD^\top \tbD \otimes \bI_P + \lambda \bGamma\right)^{-1} vec \left( \left( \bX - \bD \bA \right)^\top \tbE_{mn} \right), \\
V_{mn} &=  \bg^\top\left( \tbD^\top \tbD \otimes \bI_P + \lambda \bGamma\right)^{-1} vec \left( \tbA^\top \tbE_{mn}^\top \tbD \right),
\end{align*}
where  $\tbE_{mn}\in \bbR^{M\times N_\Lambda}$ is the indicator matrix that element $(m, n)$ of $\tbE_{mn}$ is $1$ and all other elements are zero.  

 Consider the simplification for $\bU$ first
\begin{align}
U_{mn} &= \bg^\top\left( \tbD^\top \tbD \otimes \bI_P + \lambda \bGamma\right)^{-1} \left( \tbE_{mn}^\top \otimes \bI_P \right) vec \left( \left( \bX - \bD \bA \right)^\top \right) \notag \\
&= \bg^\top\bF^{\tilde{n}} vec \left( \left( \bX - \bD \bA \right)^\top \right)_{\tilde{m}},
\end{align}
where $\bF = \left( \tbD^\top \tbD \otimes \bI_P + \lambda \bGamma\right)^{-1}$; $\tilde{m}(m) = \{ (m-1)P+1, \dots, mP \}$, $\tilde{n}(n) = \{ (n-1)P+1, \dots, nP \}$ denote the index sets;  $\bF^{\tilde{n}}$ are the $\tilde{n}$ columns of $\bF$.

Let $\bxi_m = vec \left( \left( \bX - \bD \bA \right)^\top \right)_{\tilde{m}}$. It can be shown that $\bxi_m^\top$ is the $m^{\text{th}}$ row of $(\bX - \bD \bA)$. Now the $(m, n)$ element $U_{mn}$ of the first part $\bU$ can be written as
\begin{equation}
U_{mn} =   \left(\bg^\top \bF\right)^{\tilde{n}} \bxi_{m},
\end{equation}
Stacking all elements of $\bU$
\begin{align}
\bU &= 
\begin{bmatrix}
\left(\bg^\top \bF\right)^{\tilde{1}}\bxi_{1}  &\cdots &\left(\bg^\top \bF\right)^{\widetilde{N_\Lambda}}\bxi_{1} \\
\vdots  & \ddots &\vdots \\
\left(\bg^\top \bF\right)^{\tilde{1}}\bxi_{M}  &\cdots &\left(\bg^\top \bF\right)^{\widetilde{N_\Lambda}}\bxi_{M} \\
\end{bmatrix} \notag
\\ 
& = \bxi
\begin{bmatrix}
(\bg^\top\bF)^{{\tilde{1}}^\top} \cdots  (\bg^\top\bF)^{\tilde{{N_\Lambda}}^\top}
\end{bmatrix},
\label{eq:js_v1}
\end{align}
where $\Lambda_n$ denotes the $n^\text{th}$ element of set $\Lambda$.

Now consider simplification for $\bV$. Each element $V_{mn}$ of $\bV$ can be written as
\begin{align}
V_{mn} &= \bg^\top\bF \cdot vec \left( \tbA^\top \tbE_{mn}^\top \tbD \right) \notag \\
& = \bg^\top\bF \left( \tbD^\top \tbE_{mn} \otimes \bI_P \right) vec \left( \tbA^\top \right) \notag \\
& = \bg^\top\bF 
\left(\tbD_m^\top\otimes \bI_P\right)
\tbA_n^\top,
\end{align}
where $\tbA_n$ is the $n^{\text{th}}$ row of $\tbA$ and $\tbD_m$ is the $m^{\text{th}}$ row of $\bD$.\\
Stacking every element of $\bV$, such that
\allowdisplaybreaks
\begin{align}
&\bV = 
\begin{bmatrix}
\bg^\top \bF\left(\tbD_1^\top\otimes \bI_P\right)  &\cdots &\bg^\top \bF\left(\tbD_1^\top\otimes \bI_P\right) \\
\vdots & \ddots &\vdots \\
\bg^\top \bF\left(\tbD_M^\top\otimes \bI_P\right)  &\cdots &\bg^\top \bF\left(\tbD_M^\top\otimes \bI_P\right)
\end{bmatrix} \bA^\top \notag \\
=& 
\begin{bmatrix}
\sum_{n=1}^N \bD^\top_{1n}\left( \bg^\top \bF \right)_{\bar{n}}\bA^\top_1 &\cdots &\sum_{n=1}^N \bD^\top_{1n}\left( \bg^\top \bF \right)_{\bar{n}}\bA^\top_N \\
\vdots  &\ddots &\vdots \\
\sum_{n=1}^N \bD^\top_{Mn}\left( \bg^\top \bF \right)_{\bar{n}}\bA^\top_1 &\cdots &\sum_{n=1}^N \bD^\top_{Mn}\left( \bg^\top \bF \right)_{\bar{n}}\bA^\top_N 
\end{bmatrix} \notag \\
\allowdisplaybreaks[4]
&= \bD 
\begin{bmatrix}
\left( \bg^\top \bF \right)^\top_{\bar{1}} \dots  \left( \bg^\top \bF \right)^\top_{\bar{P}}
\end{bmatrix} \bA^\top ,
\label{eq:js_v2}
\end{align}
where $\bar{p}(p) = \{ p, p+P, \dots, p + (N-1)P \}$. Combining Eq. (\ref{eq:js_v1}) and Eq. (\ref{eq:js_v2})
\begin{align}
\frac{\partial \calL}{\partial \tbD} =\bU - \bV =  \bxi \bbeta_\Lambda^\top - \tbD \bbeta_\Lambda \tbA^\top \mbox{ and } \frac{\partial \calL}{\partial \bD} =  \bxi \bbeta^\top - \bD \bbeta \bA^\top,
\end{align}
where $\bbeta_{\Lambda_c} = \mathbf{0}$ and $\bbeta_\Lambda = [\left(\bg^\top \bF \right)_{\tilde{1}}^\top, \cdots, \left(\bg^\top \bF \right)_{\overset{\sim}{N_\Lambda}}^\top]^\top$. More generally, we have defined $\bbeta_\Lambda\in \bbR^{N_\Lambda\times P}$ such that $vec(\bbeta_\Lambda^\top) = \bF \bg$.

\section{Appendix B}
\label{appendb}
The gradient for updating the dictionary can be written as 
\begin{align}
\frac{\partial \calL}{\partial D_{mn}}  & =  vec\left( \frac{\partial \calL}{\partial \bA} \right)^\top \cdot vec\left( \frac{\partial \bA}{\partial D_{mn}} \right) \notag \\
&= vec\left( \frac{\partial \calL}{\partial \bA} \right)^\top_{\Lambda} \cdot vec\left( \frac{\partial \bA}{\partial D_{mn}} \right)_{\Lambda},
\label{eq:append_eq_1}
\end{align}
Expand Eq. (\ref{eq:diff_lp_5}) and combine it with Eq. (\ref{eq:append_eq_1}), the desired gradient is
\begin{equation}
\frac{\partial \calL}{\partial D_{mn}} = U_{mn} - V_{mn} \mbox{ and } \frac{\partial \calL}{\partial \bD} = \bU - \bV,
\end{equation}
where
\begin{align*}
U_{mn} & = \bg^\top\bF^{-1}  vec \left( \bE_{mn}^\top \left( \bX - \bD \bA \right) \right)_\Lambda , \\
V_{mn} &=  \bg^\top\bF^{-1} vec \left( \bD^\top \bE_{mn}   \bA \right)_\Lambda, \\
\bF &= \left(\bI_P \otimes \bD^\top\bD + \gamma \bL\otimes \bI_N \right)_{\Lambda, \Lambda}^{-1}.
\end{align*}
Let $\bg$ has the same definition as that in  Section \ref{append}. The first part $\bU$ of $\frac{\partial f}{\partial D_{mn}}$ is
\begin{align}
U_{mn} &= \bg^\top \bF vec \left( \bE_{mn}^\top \left( \bX - \bD \bA \right) \right)_\Lambda \notag \\
&=\left( \bg^\top \bF \right)_{\tilde{n}} vec \left( \bX - \bD \bA \right)_{\tilde{m}(m,n)} 
\label{eq:v1_element}
\end{align}
 $\bE_{mn}\in\bbR^{M\times N}$ is the indicator matrix that the $(m, n)$ element is $1$ and all other elements are zero. $\tilde{m}$ and $\tilde{n}$ are defined as the following index sets,
\begin{align}
\tilde{m}(m, n) &= \{ m, \dots, m + pM, \dots \}, \forall p \mbox{ s.t. } n+pN\in \Lambda \notag \\
\tilde{n}(n) &= \{ n, n+N, \dots, n + (P-1)N \} \cap \Lambda \notag
\end{align}
%$$  and $\tilde{n}(n) = \{ n, n+N, \dots, n + (P-1)N \} \cap \Lambda$, 
Eq. (\ref{eq:v1_element}) can be further simplified by introducing $\bh^{(n)}\in\bbR^{P}$, such that
\begin{equation}
    \bh^{(n)} =
 \begin{cases}
 \left( \bg^\top \bF \right)_{n+pN}, \mbox{ if } n+pN\in \tilde{n}(n), \forall p\\
 0, \mbox{ otherwise}
  \end{cases}
\end{equation}
Now Eq. (\ref{eq:v1_element}) can be rewritten as,
\begin{equation}
U_{mn} = \bh^{(n)\top} \bxi_m,
\label{eq:appendb_simple}
\end{equation}
where $\bxi_m^\top$ is the $m^\text{th}$ row of $\bX - \bD \bA$. The first part $\bU$ of the gradient $\frac{\partial f}{\partial \bD}$ can be obtained by stacking all $U_{mn}$ in Eq. (\ref{eq:appendb_simple})
\begin{align}
\bU &= 
\begin{bmatrix}
\bxi_1^\top \bh^{(1)} &\cdots &\bx_1^\top \bh^{(N)} \\
\vdots &\ddots &\vdots\\
\bxi_M^\top \bh^{(1)} &\cdots
&\bxi_M^\top \bh^{(N)}
\end{bmatrix} \notag \\
& = \bxi \left[\bh^{(1)} \cdots \bh^{(N)} \right] \notag \\
&=\bxi \bbeta^\top,
\label{eq:lp_v1}
\end{align}
where we define $\bbeta = \left[\bh^{(1)}, \cdots, \bh^{(N)} \right]^\top \in \bbR^{N\times P}$. By examining the nonzero elements position of $\bh^{(1)}, \dots, \bh^{(N)}$, it is not difficult to find the relation between $\bbeta$ and  $\bg^\top \bF$
\begin{align}
 vec\left( \bbeta \right)_\Lambda = \bF\bg \mbox{ and } vec\left( \bbeta\right)_{\Lambda^c} =\mathbf{0}.
\end{align}
Now consider the second term $V_{mn}$ of $\frac{\partial f}{\partial D_{mn}}$ 
\begin{align}
V_{mn} &= \left( \bg^\top\bF \right)  \left( \bI_P \otimes \bD^\top\bE_{mn} \right)_{\Lambda,\Lambda} vec\left( \bA \right)_\Lambda \notag  \\
& = \left( \bg^\top\bF \right) \left(\left[ \bD_m vec(\bA)_n \dots \bD_m vec(\bA)_{(n+(P-1)N)} \right]^\top \right)_\Lambda\notag \\
& = \bD_m \sum_{p = 1}^P \bA_{n,p} \bbeta^p,
\label{eq:v2_mn_lp}
\end{align}
where $\bbeta^p$ is the $p^{\text{th}}$ column of $\bbeta$.
The differentiation $\frac{\partial f}{\partial \bD}$ can be derived from $V_{mn}$ in Eq. (\ref{eq:v2_mn_lp})
\begin{align}
\bV & = 
\begin{bmatrix}
\bD_1
\begin{bmatrix}
\sum_{p = 1}^P \bA_{1,p}\bbeta^p, \cdots, \sum_{p = 1}^P \bA_{N,p}\bbeta^p
\end{bmatrix} \\
\vdots \\
\bD_M
\begin{bmatrix}
\sum_{p = 1}^P \bA_{1,p}\left( \bF\bg \right)_{\hat{p}} \cdots \sum_{p = 1}^P \bA_{N,p}\left( \bF\bg \right)_{\hat{p}}
\end{bmatrix}
\end{bmatrix} \notag \\
&= \bD \left[  \sum_{p = 1}^P \bA_{1,p}\bbeta^p \cdots \sum_{p = 1}^P \bA_{N,p}\bbeta^p  \right] \notag \\
&= \bD \bbeta \bA^\top,
\label{eq:lp_v2}
\end{align}

Combining Eq. (\ref{eq:lp_v1}) and Eq. (\ref{eq:lp_v2}), we reach the gradient of the dictionary
\begin{align}
\frac{\partial \calL}{\partial\bD} = \bxi \bbeta^\top - \bD \bbeta \bA^\top.
\end{align}
\bibliographystyle{ieeetran}
\bibliography{refs}

\end{document}